\documentclass[10pt,twocolumn,letterpaper]{article}

\usepackage{cvpr}
\usepackage{times}
\usepackage{epsfig}
\usepackage{subfigure}
\usepackage{graphicx}
\usepackage{amsmath}
\usepackage{amssymb}

\DeclareMathOperator*{\diag}{diag}
\DeclareMathOperator*{\argmin}{arg\,min}

\newcommand{\G}{\mathcal{G}}
\newcommand{\V}{\mathcal{V}}
\newcommand{\E}{\mathcal{E}}
\newcommand{\bO}{\mathcal{O}}
\newcommand{\R}{\mathbb{R}}

\usepackage{enumitem}
\newenvironment{tight_itemize}{
\begin{itemize}[leftmargin=20pt]
  \setlength{\topsep}{0pt}
  \setlength{\itemsep}{0pt}
  \setlength{\parskip}{0pt}
  \setlength{\parsep}{0pt}
}{\end{itemize}}

\newcommand*{\affaddr}[1]{#1} 
\newcommand*{\affmark}[1][*]{\textsuperscript{#1}}
\renewcommand\thefootnote{}

\usepackage[pagebackref=true,breaklinks=true,letterpaper=true,colorlinks,bookmarks=false]{hyperref}

\cvprfinalcopy 

\def\cvprPaperID{332}

\ifcvprfinal\fi
\begin{document}

\title{Dense 3D Face Decoding over 2500FPS: Joint Texture \& Shape Convolutional Mesh Decoders}

\author{
Yuxiang Zhou \textsuperscript{*} \affmark[1] \qquad Jiankang Deng \textsuperscript{*} \affmark[1] \qquad Irene Kotsia\affmark[2] \qquad Stefanos Zafeiriou\affmark[1,3]\\
\affaddr{\affmark[1]Imperial College London} \qquad
\affaddr{\affmark[2]University of Middlesex} \qquad
\affaddr{\affmark[3]FaceSoft}\\
{\tt\small \{yuxiang.zhou10, j.deng16, s.zafeiriou\}@imperial.ac.uk}
{\tt\small , i.kotsia@mdx.ac.uk}}

\maketitle

\begin{abstract}

3D Morphable Models (3DMMs) are statistical models that represent facial texture and shape variations using a set of linear bases and more particular Principal Component Analysis (PCA). 3DMMs were used as statistical priors for reconstructing 3D faces from images by solving non-linear least square optimization problems. Recently, 3DMMs were used as generative models for training non-linear mappings (\ie, regressors) from image to the parameters of the models via Deep Convolutional Neural Networks (DCNNs). Nevertheless, all of the above methods use either fully connected layers or 2D convolutions on parametric unwrapped UV spaces leading to large networks with many parameters. In this paper, we present the first, to the best of our knowledge, non-linear 3DMMs by learning joint texture and shape auto-encoders using direct mesh convolutions. We demonstrate how these auto-encoders can be used to train very light-weight models that perform Coloured Mesh Decoding (CMD) in-the-wild at a speed of over 2500 FPS. 

\end{abstract}

\section{Introduction}

Twenty years ago, Blanz and Vetter demonstrated a remarkable achievement \cite{blanz1999morphable}. They showed that it is possible to reconstruct 3D facial geometry from a single image. This was possible by solving a non-linear optimization problem whose solution space was confined by a linear statistical model of the 3D facial shape and texture, the so-called 3D Morphable Model (3DMM). Methods based on 3DMMs are still among the state-of-the-art for 3D face reconstruction, even from images captured in-the-wild \cite{booth20163d,booth20173d,booth2018large}. 

\footnote{\textsuperscript{*}Equal contributions.} 
\setcounter{footnote}{0}
\renewcommand\thefootnote{\arabic{footnote}}

\begin{figure}[ht!]
\centering
\includegraphics[width=1\linewidth]{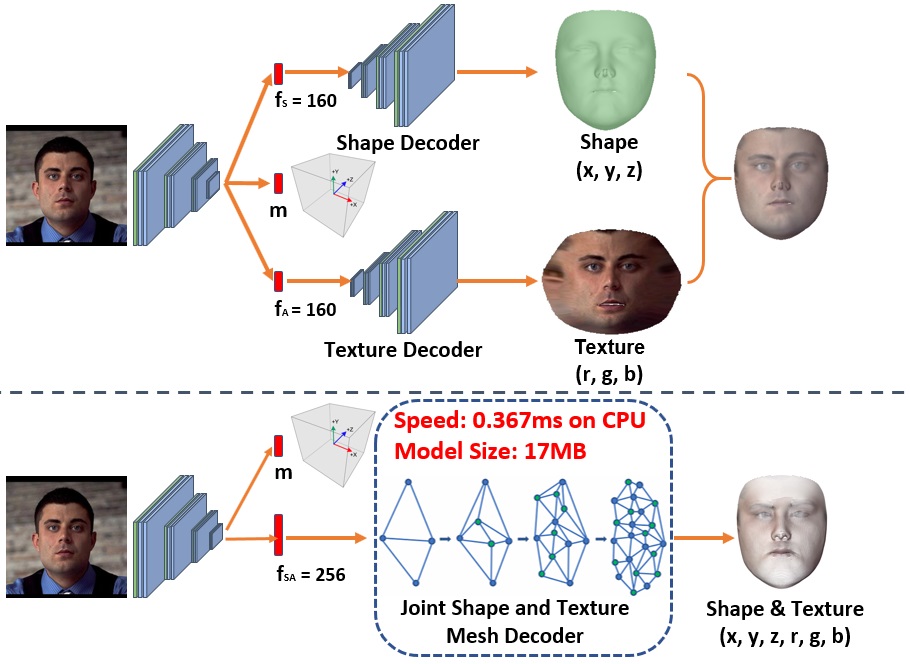}
\caption{A typical non-linear 3DMM \cite{tran2018learning} is a DCNN trained to recover shape and texture separately when given one or more 2D images. We propose a non-linear 3DMM to jointly model shape and texture by geometric convolutional networks. Our coloured mesh decoder can run over 2500 FPS with compact model size, thus being significantly faster and smaller (in terms of parameters) when compared to the PCA model.}
\label{fig:jointshapeandappearancedecoder}
\end{figure}

During the past two years, a lot of works have been conducted on how to harness the power of Deep Convolutional Neural Networks (DCNNs) for 3D shape and texture estimation from 2D facial images. The first such methods either trained regression DCNNs from image to the parameters of a 3DMM~\cite{tran2017regressing} or used a 3DMM to synthesize images and formulate an image-to-image translation problem in order to estimate the depth, using DCNNs~\cite{sela2017unrestricted}. The recent, more sophisticated, DCNN-based methods were trained using self-supervised techniques~\cite{genova2018unsupervised,tran2018nonlinear,tran2018learning} and made use of differentiable image formation architectures and differentiable renderers~\cite{genova2018unsupervised}. The most recent methods such as~\cite{tran2018nonlinear,tran2018learning}
and~\cite{tewari2017self} used self-supervision to go beyond the standard 3DMMs in terms of texture and shape. In particular,~\cite{tewari2017self} used both the 3DMMs model, as well as additional network structures (called correctives) that can capture information outside the space of 3DMMs, in order to represent the shape and texture. The method in~\cite{tran2018nonlinear,tran2018learning} tried to learn non-linear spaces (\ie, decoders, which are called non-linear 3DMMs) of shape and texture directly from the data. Nevertheless, in order to avoid poor training performance, these methods used 3DMMs fittings for the model pre-training. 

In all the above methods the 3DMMs, linear or non-linear in a form of a decoder, were modelled with either fully connected nodes~\cite{tran2017regressing} or, especially in the texture space, with 2D convolutions on unwrapped UV space~\cite{tran2018nonlinear,tran2018learning}. In this paper, we take a radically different direction. That is, motivated by the line of research on Geometric Deep Learning (GDL), a field that attempts to generalize DCNNs to non-Euclidean domains such as graphs/manifolds/meshes \cite{shuman2013emerging,defferrard2016convolutional,kipf2016semi,bronstein2017geometric,ranjan2018generating}, we make the first attempt to develop a non-linear 3DMM, that describes both shape and texture, by using mesh convolutions. Apart from being more intuitive defining non-linear 3DMMs using mesh convolutions, their major advantage is that they are defined by networks that have a very small number of parameters and hence can have very small computational complexity. In summary, the contributions of our paper are the following:

\begin{tight_itemize}
    \item We demonstrate how recent techniques that find dense or sparse correspondences (\eg, densereg \cite{guler2017densereg}, landmark localization methods \cite{zhu2016face}) can be easily extended to estimate 3D facial geometric information by means of mesh convolutional decoders. 
    \item We present the first, to the best of our knowledge, non-linear 3DMM using mesh convolutions. The proposed method decodes both shape and texture directly on the mesh domain with a compact model size ($17$MB) and amazing efficiency (over 2500 FPS on CPU). This decoder is different from the recently proposed decoder in \cite{ranjan2018generating} which only decodes 3D shape information.
    \item We propose an encoder-decoder structure that reconstructs the texture and shape directly from an in-the-wild 2D facial image. Due to the efficiency of the proposed Coloured Mesh Decoder (CMD), our method can estimate the 3D shape over $300$ FPS (for the entire system).
\end{tight_itemize}

\section{Related Work}

In the following, we briefly touch upon related topics in the literature such as linear and non-linear 3DMM representations.

\noindent{\bf Linear 3D Morphable Models.} For the past two decades, the method of choice for representing and generating 3D faces was Principal Component Analysis (PCA). PCA was used for building statistical 3D shape models (\ie, 3D Morphable Models (3DMMs)) in many works \cite{blanz1999morphable,blanz2003face,romdhani2003efficient}. Recently, PCA was adopted for building large-scale statistical models of the 3D face~\cite{booth20163d} and head~\cite{dai20173d}. It is very convenient for representing and generating faces to decouple facial identity variations from expression variations. Hence, statistical blend shape models were introduced representing only the expression variations using PCA~\cite{li2017learning,cheng20174dfab}. The original 3DMM \cite{blanz1999morphable} used a PCA model for also describing the texture variations. Nevertheless, this is quite limited in describing the texture variability in image captured in-the-wild conditions. 

\noindent{\bf Non-linear 3D Morphable Models.} In the past year, the first attempts  for learning non-linear 3DMMs were introduced \cite{tran2018nonlinear,tran2018learning,tewari2017self}.
These 3DMMs can be regarded as decoders that use DCNNs, coupled with an image-encoder. In particular, the method \cite{tewari2017self} used self-supervision to learn a new decoder with fully-connected layers that combined a linear 3DMM with new structures that can reconstruct arbitrary images. Similarly, the methods \cite{tran2018nonlinear,tran2018learning} used either fully connected layers or 2D convolutions on a UV map for decoding the shape and texture.  

All the above methods used either fully connected layers or 2D convolutions on unwrapped spaces to define the non-linear 3DMM decoders. However, these methods lead to deep networks with a large number of parameters and do not exploit the local geometry of the 3D facial structure. Therefore, decoders that use convolutions directly in the non-Euclidean facial mesh domain should be built. The field of deep learning on non-Euclidean domains, also referred to as Geometric Deep Learning~\cite{bronstein2017geometric}, has recently gained some popularity. The first works included~\cite{litany2017deformable} that proposed the so-called MeshVAE which trains a Variational-Auto-Encoder (VAE) using convolutional operators from~\cite{verma2018feastnet} and CoMA~\cite{ranjan2018generating} that used a similar architecture with spectral Chebyshev filters~\cite{defferrard2016convolutional} and additional spatial pooling to generate 3D facial meshes. The authors demonstrated that CoMA can represent better faces with expressions than PCA in a very small dimensional latent space of only eight dimensions. 

In this paper, we propose the first auto-encoder that directly uses mesh convolutions for joint texture and shape representation. This brings forth a highly effective and efficient coloured mesh decoder which can be used for 3D face reconstruction for in-the-wild data. 

\begin{figure*}[ht!]
\centering
\includegraphics[width=0.8\linewidth]{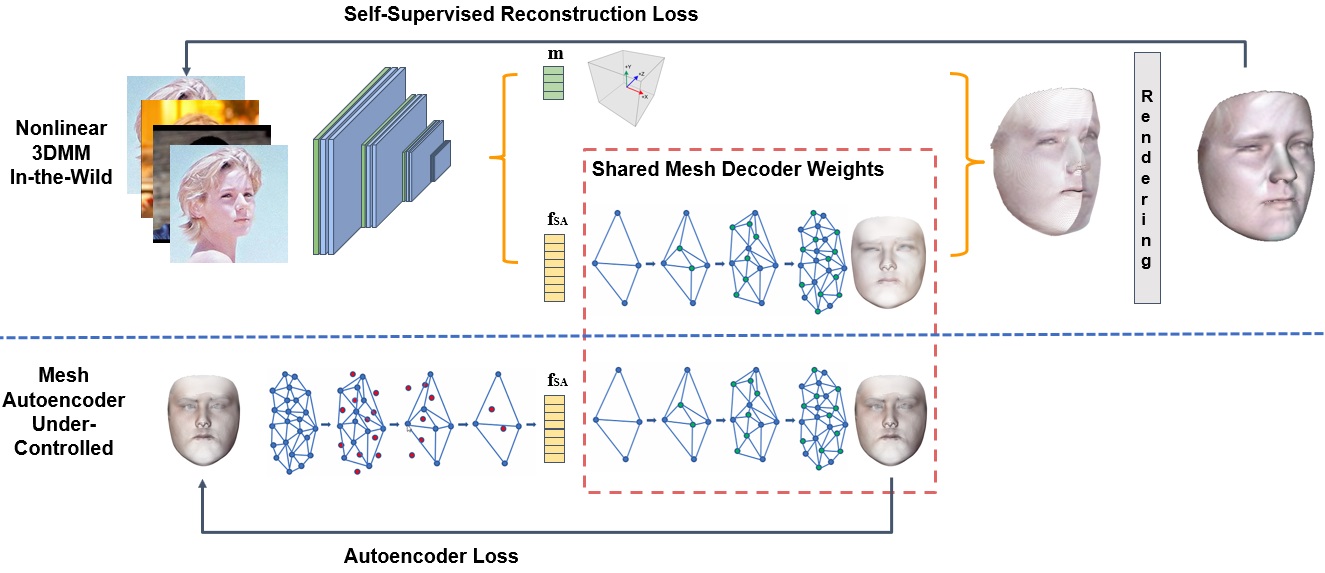}
\caption{Training procedure of the proposed method. For controlled data, we employ auto-encoder loss. For in-the-wild data, we exploit self-supervised reconstruction loss. Both models are trained end-to-end jointly with a shared coloured mesh decoder.}
\label{fig:framework}
\end{figure*}

\section{Proposed Approach}

\subsection{Coloured Mesh Auto-Encoder}

\noindent\textbf{Mesh Convolution.} We define our mesh auto-encoder based on the un-directed and connected graphs $\G=(\V,\E)$, where $\V \in \mathbb{R}^{n \times 6}$ is a set of $n$ vertices containing the joint shape (\eg x, y, z) and texture (\eg r, g, b) information, and $\E \in \{0,1\}^{n \times n}$ is an adjacency matrix encoding the connection status between vertices. 

Following~\cite{defferrard2016convolutional,ranjan2018convolutional}, the non-normalized graph Laplacian is defined as $L = D -\E \in\R^{n \times n}$ where $D \in \R^{n \times n}$ is the diagonal matrix with $D_{ii} = \sum_j \E_{ij}$ and the normalized definition is $L = I_n -D^{-1/2} \E D^{-1/2}$ where $I_n$ is the identity matrix. The Laplacian $L$ can be diagonalized by the Fourier bases $U=[u_0, \ldots, u_{n-1}] \in \R^{n \times n}$ such that $L = U \Lambda U^T$ where $\Lambda = \diag([\lambda_0, \ldots, \lambda_{n-1}]) \in \R^{n \times n}$. The graph Fourier transform of our face representation $x \in \R^{n \times 6}$ is then defined as $\hat{x} = U^T x$, and its inverse as $x = U \hat{x}$.

The operation of the convolution on a graph can be defined by formulating mesh filtering with a kernel $g_\theta$ using a recursive Chebyshev polynomial~\cite{defferrard2016convolutional,ranjan2018convolutional}. The filter $g_\theta$ can be parameterized as a truncated Chebyshev polynomial expansion of order $K$,
\begin{equation}
g_\theta(\Lambda) = \sum_{k=0}^{K-1} \theta_k T_k(\tilde{\Lambda}),
\label{chebyshevpolynomial}
\end{equation}
where $\theta \in \R^K$ is a vector of Chebyshev coefficients and $T_k(\tilde{\Lambda}) \in \R^{n \times n}$ is the Chebyshev polynomial of order
$k$ evaluated at a scaled Laplacian $\tilde{\Lambda} = 2 \Lambda / \lambda_{max} - I_n$. $T_k$ can be recursively computed by $T_k(x) = 2xT_{k-1}(x) - T_{k-2}(x)$ with $T_0=1$ and $T_1=x$. 

The spectral convolution can be defined as
\begin{equation}
y_j = \sum_{i=1}^{F_{in}} g_{\theta_{i,j}}(L) x_{i},
\label{spectralconvolution}
\end{equation}
where $x \in \mathbb{R}^{n \times F_{in}}$ is the input and $y \in \mathbb{R}^{n \times F_{out}}$ is the output. The entire filtering operation $y = g_\theta(L) x$ is very efficient and only costs $\bO(K|\E|)$ operations.

\noindent\textbf{Mesh Down-sampling and Up-sampling.} We follow~\cite{ranjan2018convolutional} to employ a binary transformation matrix $Q_d \in \{0,1\}^{n \times m}$ to perform down-sampling of a mesh with $m$ vertices and conduct up-sampling using another transformation matrix $Q_u \in \mathbb{R}^{m \times n}$. 

$Q_d$ is calculated by iteratively contracting vertex pairs under the constraint of minimizing quadric error \cite{garland1997surface}.
During down-sampling, we store the barycentric coordinates of the discarded vertices with regard to the down-sampled mesh so that the up-sampling step can add new vertices with the same barycentric locations information.

For up-sampling, vertices directly retained during the down-sampling step undergo convolutional transformations. Vertices discarded during down-sampling are mapped into the down-sampled mesh surface using recorded barycentric coordinates. The up-sampled mesh with vertices $\V_u$ is efficiently predicted by a sparse matrix multiplication, $\V_u = Q_u \V_d$.

\subsection{Coloured Mesh Decoder in-the-Wild}

The non-linear 3DMM fitting in-the-wild is designed in an unsupervised/self-supervised manner. As we are able to construct joint shape \& texture bases with the coloured mesh auto-encoder, the problem can be treated as a matrix multiplication between the bases and the optimal coefficients that reconstruct the 3D face. From the perspective of a neural network, this can be viewed as an image encoder $E_{I}(I;\mathbf{\theta_{I}})$ that is trained to regress to the 3D shape and texture, noted as $f_{SA}$. As shown in Fig.~\ref{fig:framework}, a 2D convolution network is used to encode in-the-wild images followed by a mesh decoder $\mathcal{D}(\mathbf{f_{SA}};\mathbf{\theta_D})$, whose weights are shared across the decoder~\cite{chrysos2018robust} in the mesh auto-encoder. However, the output of the joint shape \& texture decoder is a coloured mesh within a unit sphere. Like linear 3DMM~\cite{booth20173d}, a camera model is required to project the 3D mesh from the object-centered Cartesian coordinates into an image plane in the same Cartesian coordinates.

\noindent\textbf{Projection Model.} We employ a pinhole camera model in this work, which utilizes a perspective transformation model. The parameters of the projection operation can be formulated as following:
\begin{equation}
\mathbf{c} = [p_x,p_y,p_z,o_x,o_y,o_z,u_x,u_y,u_z,f]^\mathsf{T},
\label{eq:camerapose}
\end{equation}
where $\mathbf{p},\mathbf{o},\mathbf{u}$ represent camera position, orientation and upright direction, respectively, in Cartesian coordinates. $f$ is the field of view (FOV) that controls the perspective projection. We also concatenate lighting parameters together with camera parameters as rendering parameters that will be predicted by the image encoder. Three point light sources and constant ambient light are assumed, to a total of 12 parameters $\mathbf{l}$ for lighting. For abbreviation, we represent the rendering parameter $\mathbf{m}=[\mathbf{c}^T, \mathbf{l}^T]^T$ as a vector of size 22 and the projection model as the function $\mathbf{{\hat{I}}}=\mathcal{P}(\mathcal{D}(\mathbf{f_{SA}});\mathbf{m}): \mathbb{R}^{3N} \rightarrow \mathbb{R}^{2N}$.

\noindent\textbf{Differentiable Renderer.} To make the network end-to-end trainable, we incorporated a differentiable renderer~\cite{genova2018unsupervised} to project the output mesh $\mathcal{D}(\mathbf{f_{SA}})$ onto the image plane $\mathbf{{\hat{I}}}$. The $l_1$ norm is pixel-wisely calculated as the loss function. The renderer, also known as rasterizer, generates barycentric coordinates and corresponding triangle IDs for each pixel at the image plane. The rendering procedure involves Phong shading~\cite{phong1975illumination} and interpolating according to the barycentric coordinates. Also, camera and illumination parameters are computed in the same framework. The whole pipeline is able to be trained end-to-end with the loss gradients back-propagated through the differentiable renderer. 

\noindent\textbf{Losses.} We have formulated a loss function applied jointly to under-controlled coloured mesh auto-encoder and in-the-wild coloured mesh decoder, thus enabling supervised and self-supervised end-to-end training. It is formulated as below:
\begin{align}
\argmin_{\mathbf{{\theta_{E_{M}}}},\mathbf{\theta_{E_{I}}},\mathbf{\theta_{D}}, \mathbf{m}} L_{rec}+\lambda L_{render}.
\label{eq:loss}
\end{align}
Where the objective function:
\begin{align}
L_{\text{rec}} &= \sum_{i}||\mathcal{D}(E_{M}(\mathbf{S}_i;\mathbf{\theta_{E_{M}}});\mathbf{\theta_{D}}) - \mathbf{S}_i||_2 \nonumber \\
               &+\sum_{i}||\mathcal{D}(E_{M}(\mathbf{A}_i;\mathbf{\theta_{E_{M}}});\mathbf{\theta_{D}}) - \mathbf{A}_i||_1
\label{eq:undercontrolled}
\end{align}
is applied to enforce shape and texture reconstruction of the coloured mesh auto-encoder, in which $l_2$ and $l_1$ norms are applied on shape $S$ and texture $A$, respectively. The term:
\begin{align}
L_{render} = \sum_{i}||\mathcal{P}(\mathcal{D}(E_{I}(\mathbf{I}_i;\mathbf{\theta_{E_{I}}});\mathbf{\theta_{D}});\mathbf{m}) - \mathbf{I}_i||_1
\label{eq:inthewild}
\end{align}
represents the pixel-wise reconstruction error for in-the-wild images when applying a mask to only visible facial pixels. We use $\lambda=0.01$ and gradually increase to $1.0$ during training. 

\section{Experimental Results}

\subsection{Datasets}

We train our method using both under-controlled data (3DMD~\cite{deng2017uv}) and in-the-wild data (300W-LP~\cite{zhu2016face} and CelebA~\cite{liu2015deep}).
The 3DMD dataset~\cite{deng2017uv} contains around $21$k raw scans of 3,564 unique identities with expression variations.
The 300W-LP dataset~\cite{zhu2016face} consists of about $60$k large pose facial data, which are 
synthetically generated by the profiling method of~\cite{zhu2016face}.
The CelebA dataset~\cite{liu2015deep} is a large-scale face attributes dataset with more than $200$k celebrity images, which cover large pose variations and background clutter. 
Each training image is cropped to bounding boxes of indexed 68 facial landmarks with random perturbation to simulate a coarse face detector. 

We perform extensive qualitative experiments on AFLW2000-3D~\cite{zhu2016face}, 300VW~\cite{sagonas2013semi} and CelebA testset~\cite{liu2015deep}. We also conducted quantitative comparisons with prior works on FaceWarehouse~\cite{cao2014facewarehouse} and Florence~\cite{bagdanov2011florence}, where accurate 3D meshes are available for evaluation. FaceWarehouse is a 3D facial expressions database collected by a Kinect RGBD camera. 150 candidates aged from 7 to 80 of various ethnic groups are involved. Florence is a 3D face dataset that contains 53 subjects with their ground truth 3D meshes acquired from a structured-light scanning system.

\subsection{Implementation Details}

\noindent\textbf{Network Architecture.} Our architecture consists of four sub-modules as shown in Fig.~\ref{fig:framework}, named Image Encoder~\cite{tran2018nonlinear,tran2018learning}, Coloured Mesh Encoder~\cite{ranjan2018convolutional}, a shared Coloured Mesh Decoder~\cite{ranjan2018convolutional} and a differentiable rendering module~\cite{genova2018unsupervised}. The image encoder part takes input images of shape $112\times112\times3$ followed by 10 convolution layers. It reduces the dimension of the input images to $7\times7\times256$ and applies a fully connected layer that constructs a $256\times1$-dimension embedding space. Every convolutional layer is followed by a batch normalization layer and a ReLU activation layer. The kernel size of all convolution layers is 3 and the stride is 2 for any down-sampling convolution layer. The coloured mesh decoder takes an embedding of size $256\times1$ and decodes to a coloured mesh of size $28431 \times 6$ (3 shape and 3 texture channels). The encoder/decoder consists of 4 geometric convolutional filters~\cite{ranjan2018convolutional}, each one of which is followed by a down/up-sampling layer that reduces/increases the number of vertices by 4 times. Every graph convolutional layer is followed by a ReLU activation function similar to those in the image encoder. 

\noindent\textbf{Training Details.} Both (1) the under-controlled coloured mesh auto-encoder and (2) the in-the-wild coloured mesh decoder are jointly trained end-to-end although each one uses a different data source. Both models are trained with Adam optimizer with a start learning rate of 1e-4. A learning rate decay is applied with the rate at 0.98 of each epoch. We train the model for 200 epochs. We perturb the training image with a random flipping, random rotation, random scaling and random cropping to the size of $112\times112$ from a $136\times136$ input. 

\subsection{Ablation Study on Coloured Mesh Auto-Encoder}

\noindent\textbf{Reconstruction Capacity.} We compare the power of linear and non-linear 3DMMs in representing real-world 3D scans with different embedding dimensions to emphasize the compactness of our coloured mesh decoder. Here, we use $10\%$ of 3D face scans from the 3DMD dataset as the test set. 

\begin{figure}[h]
\centering
\begin{minipage}{1.0\columnwidth}
\begin{center}
\begin{tabular}{ @{}c@{}c@{}c@{}c@{}c}
    3D Scan & \multicolumn{3}{c}{Coloured Mesh Decoder}  & Linear \\
    \hline
            & $64$ & $128$ & $256$ & $178$ \\
    \includegraphics[trim = 128 104 126 122, clip, width=0.16\textwidth]{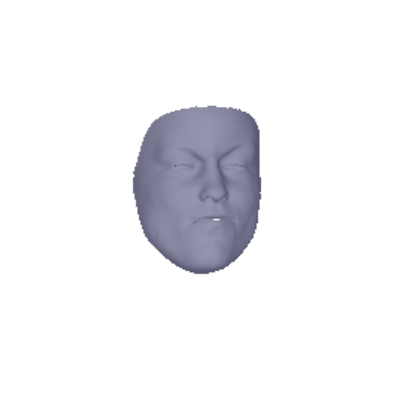} &
    \includegraphics[trim = 128 104 126 122, clip, width=0.16\textwidth]{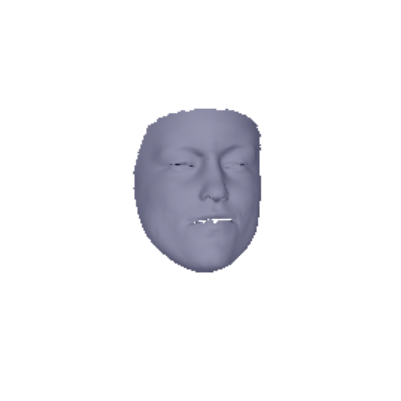} &
    \includegraphics[trim = 128 104 126 122, clip, width=0.16\textwidth]{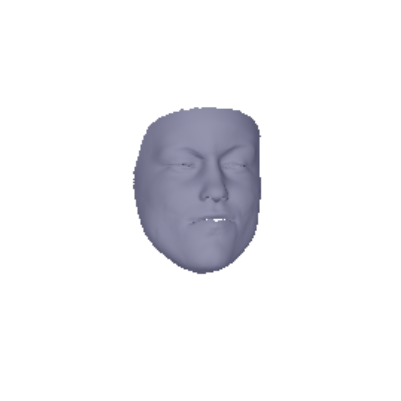} &
    \includegraphics[trim = 128 104 126 122, clip, width=0.16\textwidth]{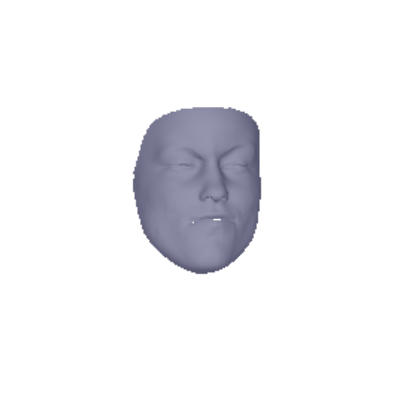} &
    \includegraphics[trim = 128 104 126 122, clip, width=0.16\textwidth]{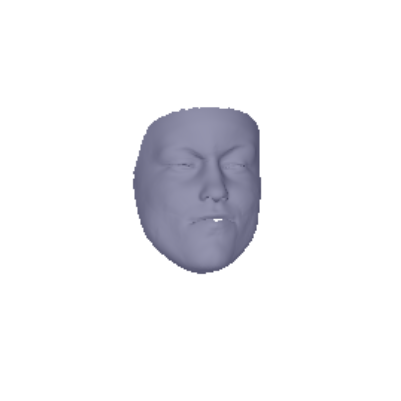} \\
    
    \includegraphics[trim = 242 168 193 181, clip, width=0.16\textwidth]{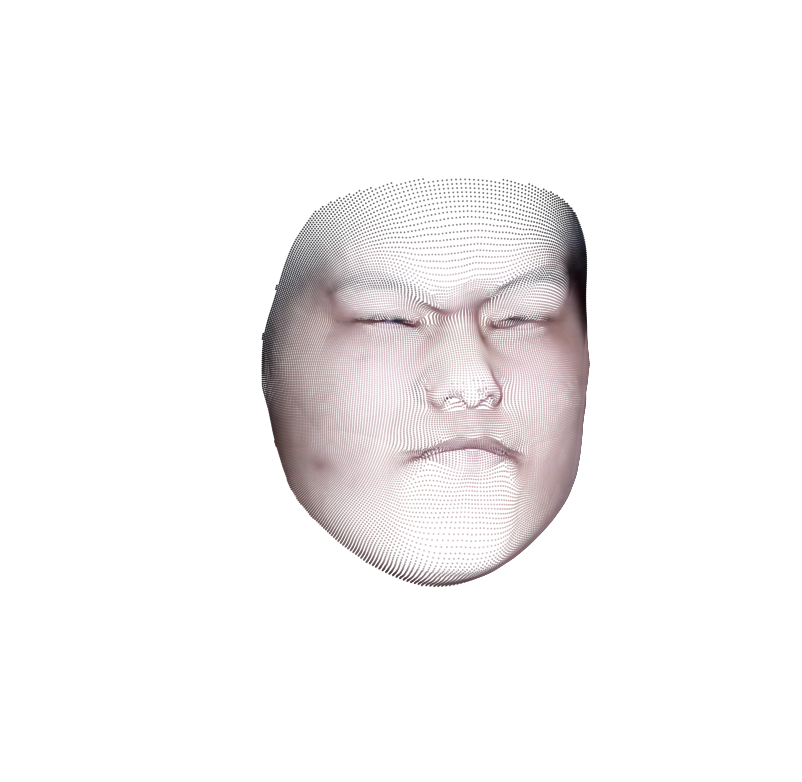} &
    \includegraphics[trim = 242 168 193 181, clip, width=0.16\textwidth]{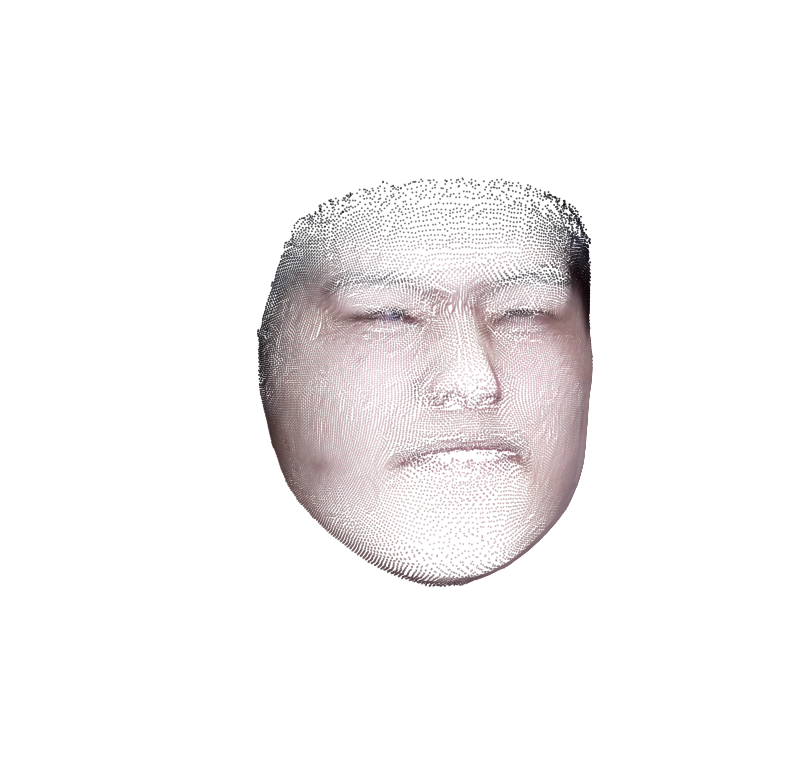} &
    \includegraphics[trim = 242 168 193 181, clip, width=0.16\textwidth]{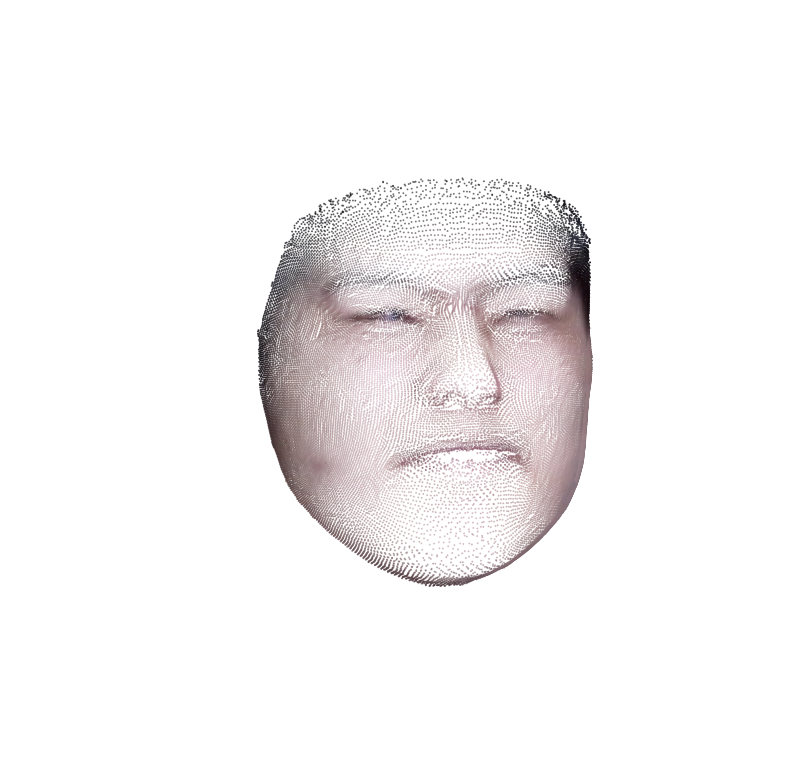} &
    \includegraphics[trim = 242 168 193 181, clip, width=0.16\textwidth]{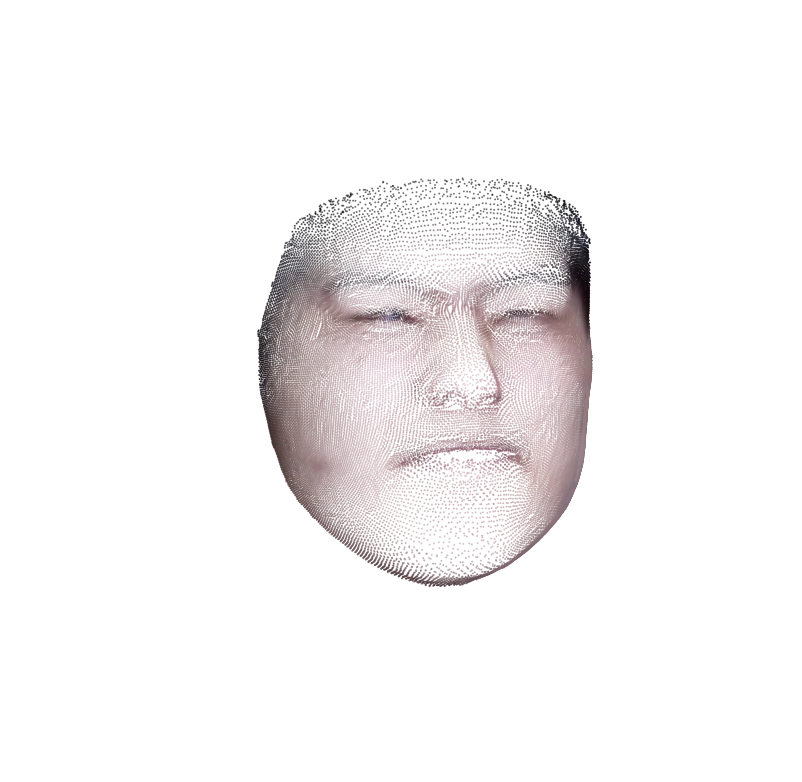} &
    \includegraphics[trim = 242 168 193 181, clip, width=0.16\textwidth]{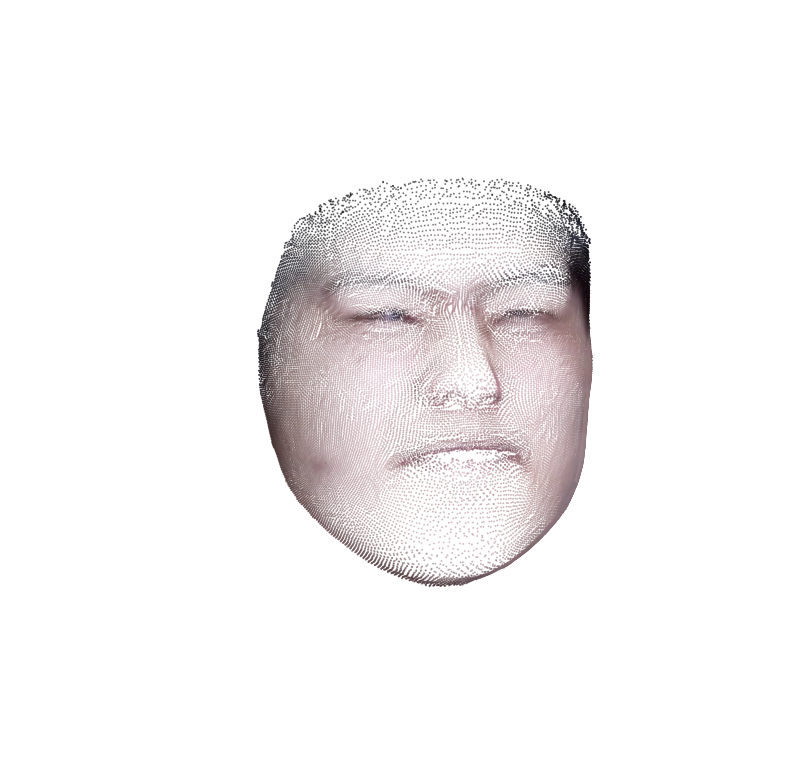} \\
    \hline
    \multicolumn{5}{c}{Expression Embedding}     \\
    \hline
    \includegraphics[width=0.16\textwidth]{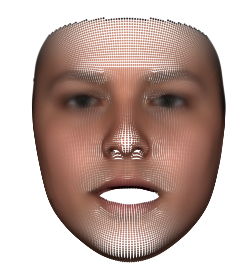} &
    \includegraphics[width=0.16\textwidth]{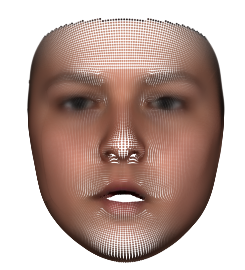} &
    \includegraphics[width=0.16\textwidth]{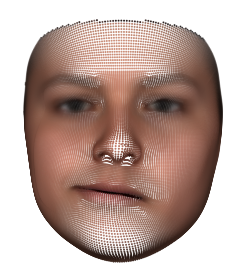} &
    \includegraphics[width=0.16\textwidth]{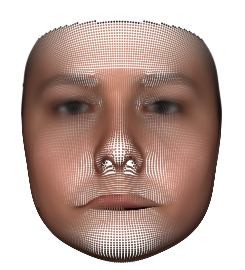} &
    \includegraphics[width=0.16\textwidth]{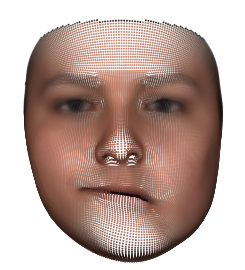} \\
    
    \includegraphics[width=0.16\textwidth]{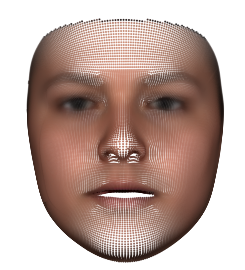} &
    \includegraphics[width=0.16\textwidth]{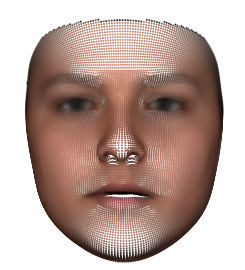} &
    \includegraphics[width=0.16\textwidth]{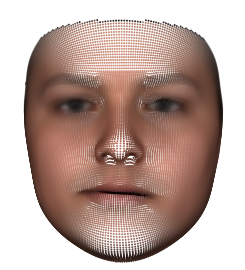} &
    \includegraphics[width=0.16\textwidth]{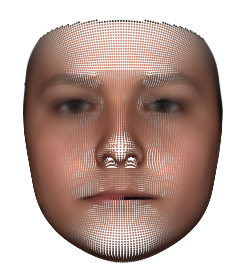} &
    \includegraphics[width=0.16\textwidth]{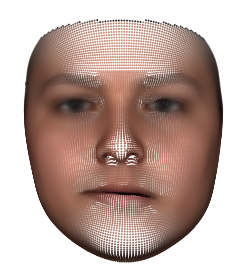} \\
    
    \includegraphics[width=0.16\textwidth]{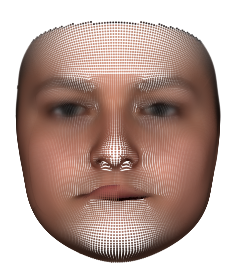} &
    \includegraphics[width=0.16\textwidth]{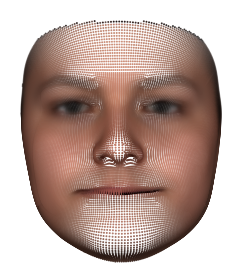} &
    \includegraphics[width=0.16\textwidth]{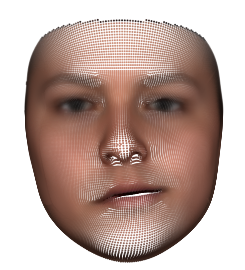} &
    \includegraphics[width=0.16\textwidth]{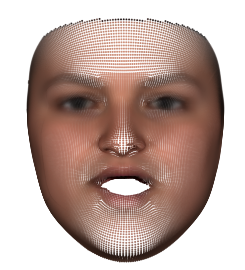} &
    \includegraphics[width=0.16\textwidth]{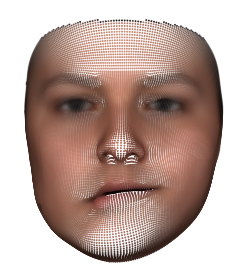} \\
    \hline
    \multicolumn{5}{c}{Illumination Embedding}     \\
    \hline
    \includegraphics[width=0.16\textwidth]{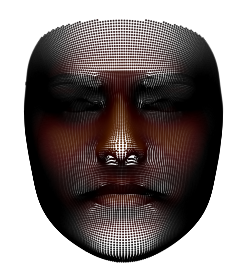} &
    \includegraphics[width=0.16\textwidth]{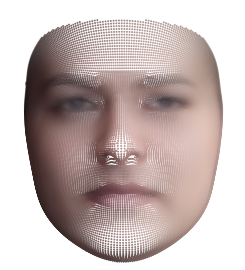} &
    \includegraphics[width=0.16\textwidth]{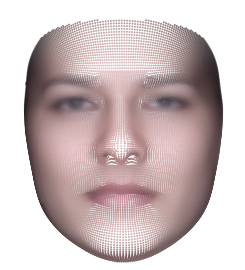} &
    \includegraphics[width=0.16\textwidth]{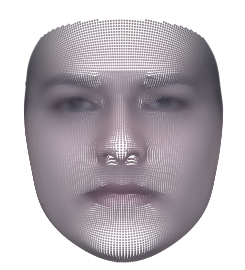} &
    \includegraphics[width=0.16\textwidth]{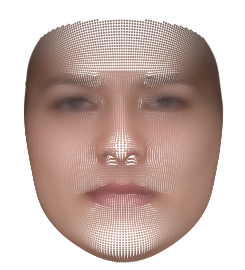} \\
    
    \includegraphics[width=0.16\textwidth]{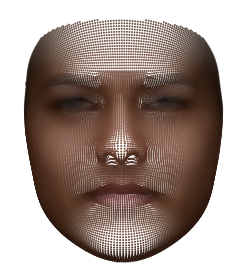} &
    \includegraphics[width=0.16\textwidth]{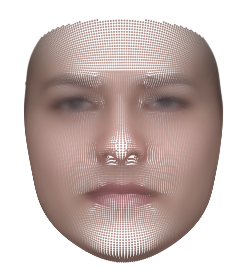} &
    \includegraphics[width=0.16\textwidth]{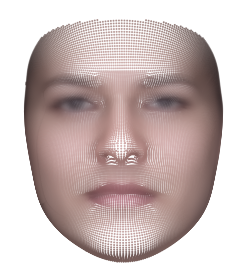} &
    \includegraphics[width=0.16\textwidth]{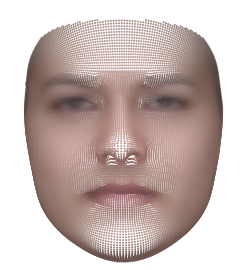} &
    \includegraphics[width=0.16\textwidth]{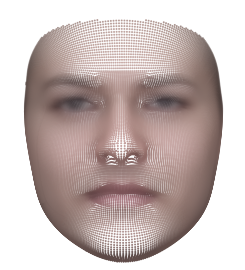} \\
    
    \includegraphics[width=0.16\textwidth]{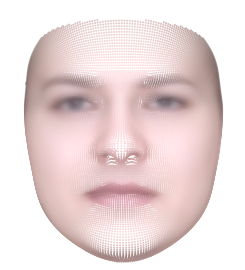} &
    \includegraphics[width=0.16\textwidth]{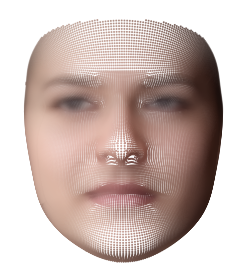} &
    \includegraphics[width=0.16\textwidth]{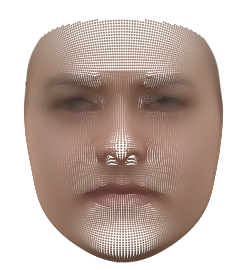} &
    \includegraphics[width=0.16\textwidth]{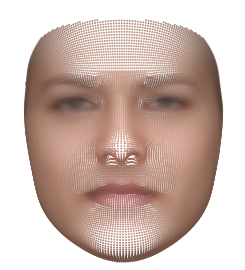} &
    \includegraphics[width=0.16\textwidth]{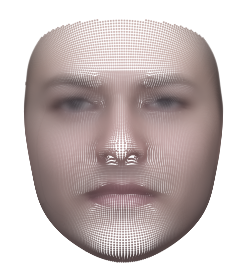} \\
    \hline
    \multicolumn{5}{c}{Beard Embedding} \\
    \hline
    \includegraphics[width=0.18\textwidth]{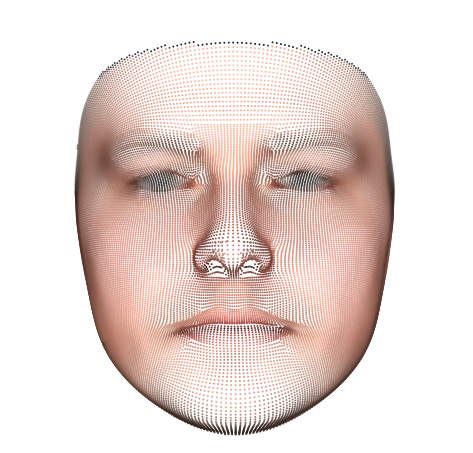} &
    \includegraphics[width=0.18\textwidth]{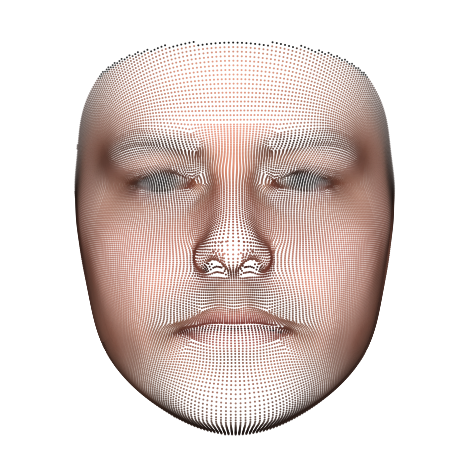} &
    \includegraphics[width=0.18\textwidth]{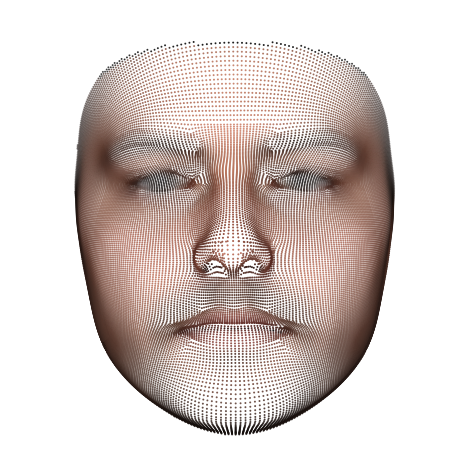} &
    \includegraphics[width=0.18\textwidth]{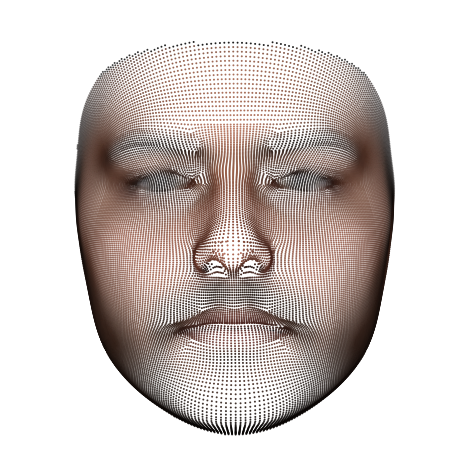} &
    \includegraphics[width=0.18\textwidth]{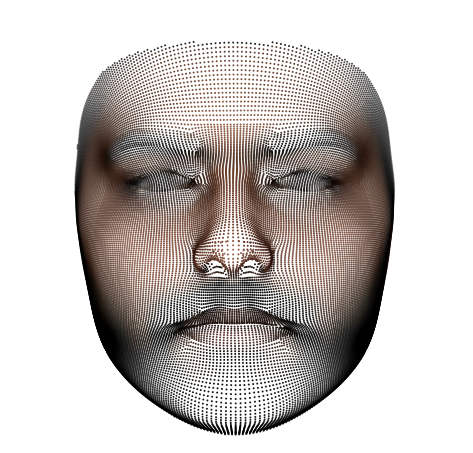} \\
    
    \hline
\end{tabular}
\end{center}
\caption{Shape and texture representations followed by expression, illumination and beard embedding generated by the proposed coloured mesh decoder.}
\label{fig:representation}
\end{minipage}
\end{figure}

As illustrated in the top of Fig.~\ref{fig:representation}, we compare the visual quality of reconstruction results produced by linear and non-linear models. To quantify the results of shape modelling, we use the Normalized Mean Error (NME), which is the averaged per-vertex errors between the ground-truth shapes and the reconstructed shapes normalized by inter-ocular distances. For evaluation of texture modelling, we employ the pixel-wise Mean Absolute Error (MAE) between the ground-truth and reconstructed texture. 

As shown in Tab.~\ref{tab:shapeandappearancerepresentation},
our non-linear shape model has a significantly smaller shape reconstruction error than the linear model. Moreover, the joint non-linear model notably reduces the reconstruction error even further, indicating that integrating texture information is helpful to constrain the deformation of vertices. For the comparison on the texture reconstruction, a slightly higher reconstruction error of texture is expected as the missing texture information between vertices was interpolated in our model, while a linear model has the full texture information.

\begin{table}[t!]
\small
\begin{center}
\begin{tabular}{c|c|c}
\hline
$Dimension$ & Shape &  Texture   \\ 
\hline
PCA $f_{S/A}$=64    & $0.0313$ & $0.0196$ \\
PCA $f_{S/A}$=128   & $0.0280$ & $0.0169$ \\
PCA $f_{S/A}$=185   & $0.0237$ & $0.0146$ \\
\hline
$f_S$=64    & $0.0304$ & - \\
$f_S$=128   & $0.0261$ & - \\
$f_S$=256   & $0.0199$ & - \\
\hline
$f_{SA}$=64    & $0.0286$ & $0.0325$ \\
$f_{SA}$=128   & $0.0220$ & $0.0271$  \\
$f_{SA}$=256   & $\mathbf{0.0133}$ & $0.0228$ \\
\hline
\end{tabular}
\end{center}
\caption{3D scan face reconstructions comparison (NME for shape and $l_1$ channel-wise error for texture).} 
\label{tab:shapeandappearancerepresentation}
\end{table}

\noindent\textbf{Attribute Embedding.} To get a better understanding of different faces embedded in our coloured mesh decoder, we investigate the semantic attribute embedding. For a given attribute, \eg, smile, we feed the face data (shape and texture) with that attribute $\{\mathbf{I}_i\}_{i=1}^n$ into our coloured mesh encoder to obtain the embedding parameters $\{\mathbf{f}_{SA}^{i}\}_{i=1}^n$, which represent corresponding distributions of the attribute in the low dimensional embedding space. Taking the mean parameters $\mathbf{\bar{f}}_{SA}$ as input to the trained coloured mesh decoder, we can reconstruct the mean shape and texture with that attribute. Based on the principal component analysis on the embedding parameters $\{\mathbf{f}_{SA}^{i}\}_{i=1}^n$, we can conveniently use one variable (principal component) to change the attribute. Fig.~\ref{fig:representation} shows some 3D shapes with texture sampled from the latent space. Here, we can observe that the power of our non-linear coloured mesh decoder is excellent at modelling expressions, illuminations and even beards with a tight embedding dimension ($f_{SA}=256$).

\subsection{Coloured Mesh Decoder Applied In-the-wild}

\subsubsection{3D Face Alignment}
\label{sec:3d_face_alignment}

Since our method can model shape and texture simultaneously, we apply it for 3D morphable fitting in the wild
and test the performance on the task of sparse 3D face alignment. We compare our model with the most recent state-of-the-art methods, \eg 3DDFA~\cite{zhu2016face}, N-3DMM~\cite{tran2018nonlinear} and PRNet~\cite{feng2018joint} on the AFLW2000-3D~\cite{zhu2016face} dataset. The accuracy is evaluated by the Normalized Mean Error (NME), that is the average of landmark error normalized by the bounding box size on three pose subsets~\cite{zhu2016face}.

\begin{table}[t!]
\begin{center}
\small
\begin{tabular}{c|c|c|c|c}
\hline
Method&3DDFA\cite{zhu2016face} & N3DMM \cite{tran2018learning}  & PRNet \cite{feng2018joint} & CMD  \\
\hline
NME    & 5.42     & 4.12    & 3.62  & 3.98  \\
\hline
\end{tabular}
\end{center}
\caption{Face alignment results ($\%$) on the AFLW2000-3D dataset. Performance is reported as bounding box size normalized mean error~\cite{zhu2016face}.}
\label{table:AFLW20003D}
\end{table}

3DDFA~\cite{zhu2016face} is a cascade of CNNs that iteratively refines its estimation in multiple steps. 
N-3DMM~\cite{tran2018learning} utilizes the 2D deep convolutional neural networks to build a non-linear 3DMM on the UV position and texture maps, and fits the unconstrained 2D in-the-wild face images in a weakly supervised way.
By contrast, our method employs the coloured mesh decoder to build the non-linear 3DMM.
Our model not only has better performance but also has a more compact model size and a more efficient running time.
PRNet~\cite{tran2018learning} employs an encoder-decoder neural network to directly regress the UV position map.
The performance of our method is slightly worse than PRNet majorly due to the complexity of the network. 

\begin{figure}[t!]
\begin{center}
\begin{tabular} {c@{\hskip 0.01mm}c@{\hskip 0.01mm}c@{\hskip 0.01mm}c@{\hskip 0.01mm}c@{\hskip 0.01mm}c@{\hskip 0.01mm}c}
\includegraphics[width=0.075\textwidth]{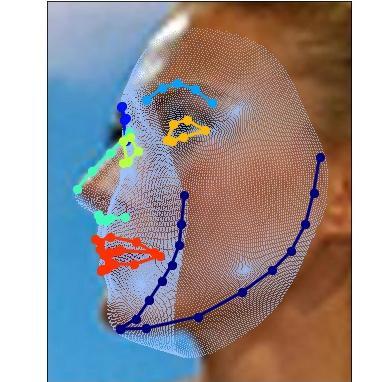} &
\includegraphics[width=0.075\textwidth]{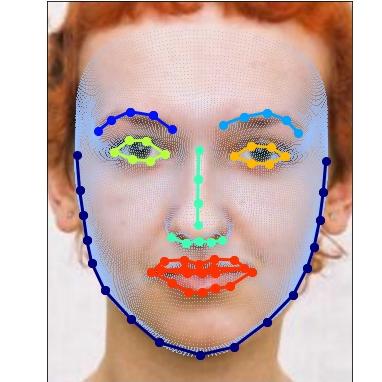} &
\includegraphics[width=0.075\textwidth]{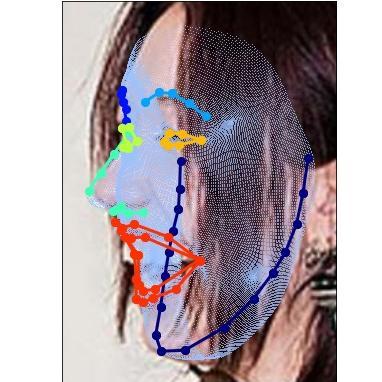} &
\includegraphics[width=0.075\textwidth]{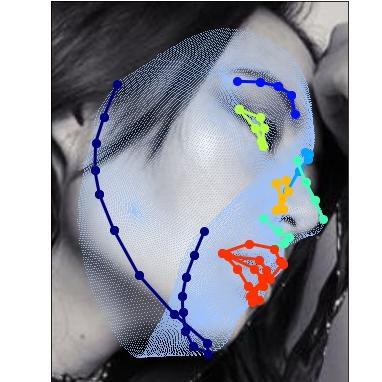} &
\includegraphics[width=0.075\textwidth]{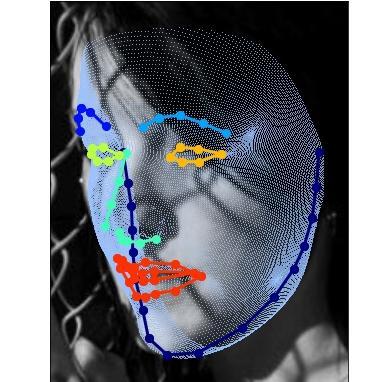} &
\includegraphics[width=0.075\textwidth]{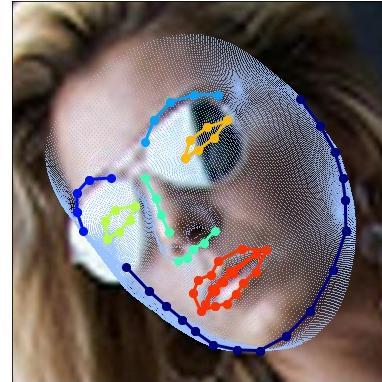} \\

\includegraphics[width=0.075\textwidth]{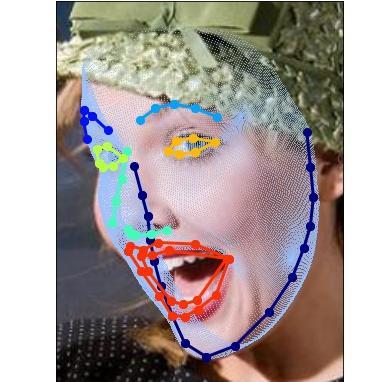} &
\includegraphics[width=0.075\textwidth]{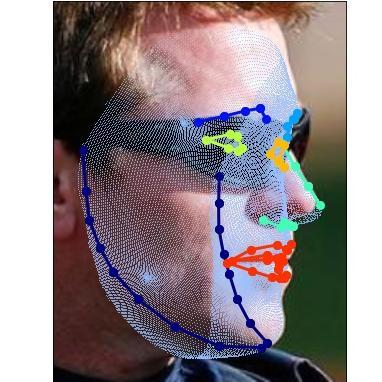} &
\includegraphics[width=0.075\textwidth]{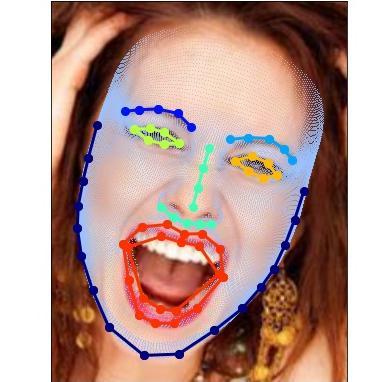} &
\includegraphics[width=0.075\textwidth]{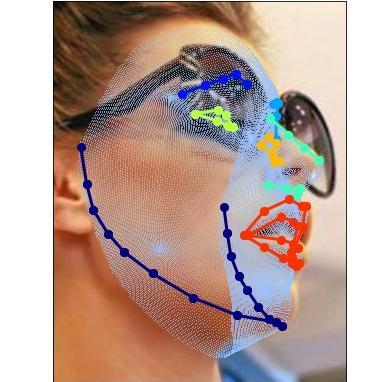} &
\includegraphics[width=0.075\textwidth]{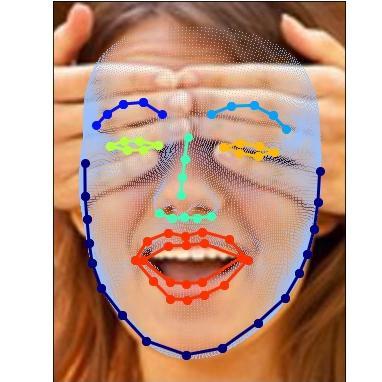} &
\includegraphics[width=0.075\textwidth]{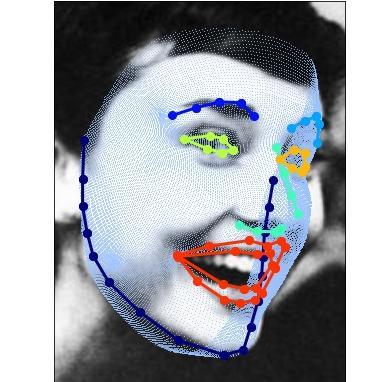} \\

\includegraphics[width=0.075\textwidth]{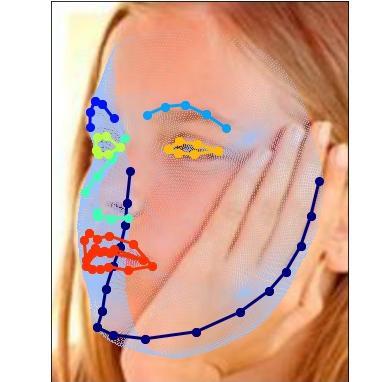} &
\includegraphics[width=0.075\textwidth]{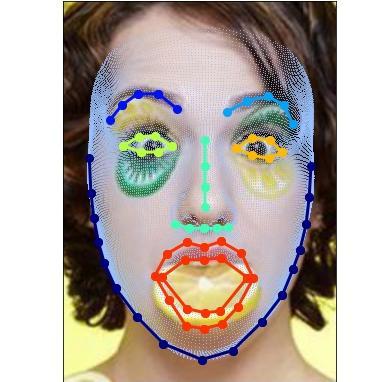} &
\includegraphics[width=0.075\textwidth]{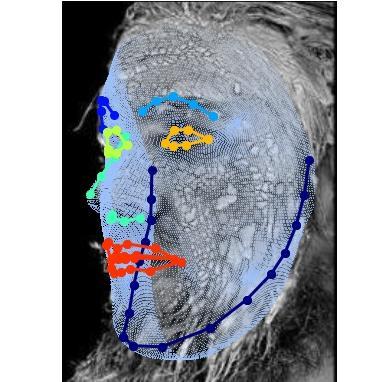} &
\includegraphics[width=0.075\textwidth]{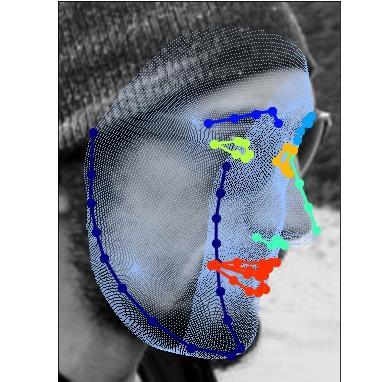} &
\includegraphics[width=0.075\textwidth]{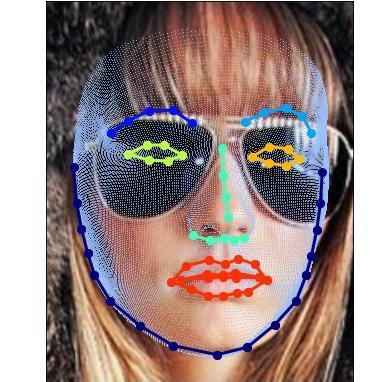} &
\includegraphics[width=0.075\textwidth]{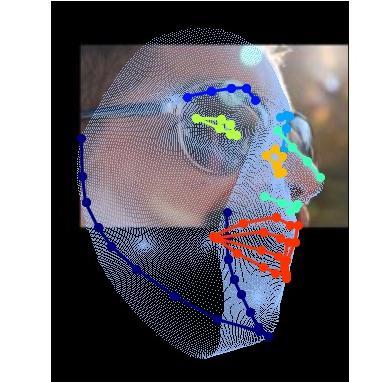} \\
\end{tabular}
\end{center}
\vspace{-2mm}
\caption{Face alignment results on the AFLW2000-3D dataset. The proposed method can handle extreme pose, expression, occlusion and illumination.}
\label{fig:facealignmentaflw20003d}
\end{figure}

In Fig.~\ref{fig:facealignmentaflw20003d}, we give some exemplary alignment results, which demonstrate successful sparse 3D face alignment results under extreme poses, exaggerated expressions, heavy occlusions and variable illuminations. We also see that the dense shape (vertices) predictions are also very robust in the wild, which means that for any kind of facial landmark configuration our method is able to give accurate localization results if the landmark correspondence with our shape configuration is given. 

\subsubsection{3D Face Reconstruction}

We first qualitatively compare our approach with five recent state-of-the-art 3D face reconstruction methods: (1) 3DMM fitting networks learned in a supervised way (Sela \etal~\cite{sela2017unrestricted}), (2) 3DMM fitting networks learned in an unsupervised way named MoFA (Tewari \etal~\cite{tewari2017mofa}), (3) a direct volumetric CNN regression approach called VRN (Jackson \etal~\cite{jackson2017large}), (4) a direct UV position map regression method named PRNet (Feng \etal~\cite{feng2018joint}), (5) a non-linear 3DMM fitting networks learned in weakly supervised fashion named N-3DMM (Tran \etal \cite{tran2018learning}). As PRNet and N-3DMM both employ 2D convolution networks on the UV position map to learn the shape model, we view PRNet and N-3DMM as the closest baselines to our method.

\begin{figure*}[t!]
\begin{center}
\small
\begin{tabular}{c@{\hskip 2mm}c@{\hskip 1mm}c@{\hskip 1mm}c@{\hskip 2mm}c@{\hskip 2mm}c@{\hskip 1mm}c@{\hskip 2mm}c@{\hskip 1mm}c@{\hskip 1mm}c@{}}

Input & \multicolumn{3}{c}{Sela \cite{sela2017unrestricted}} &  PRNet \cite{feng2018joint}  & \multicolumn{2}{c}{N-3DMM \cite{tran2018learning}} & \multicolumn{3}{c}{CMD}\\

\includegraphics[width=0.092\textwidth]{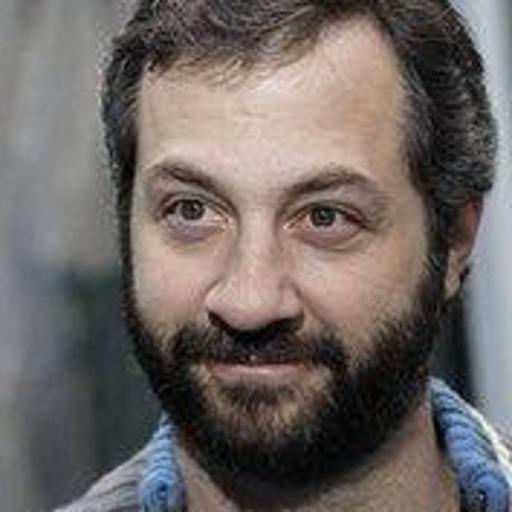} &

\includegraphics[width=0.092\textwidth]{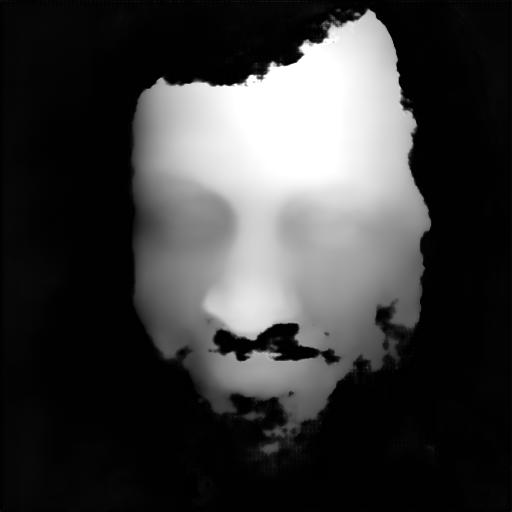} &
\includegraphics[width=0.092\textwidth]{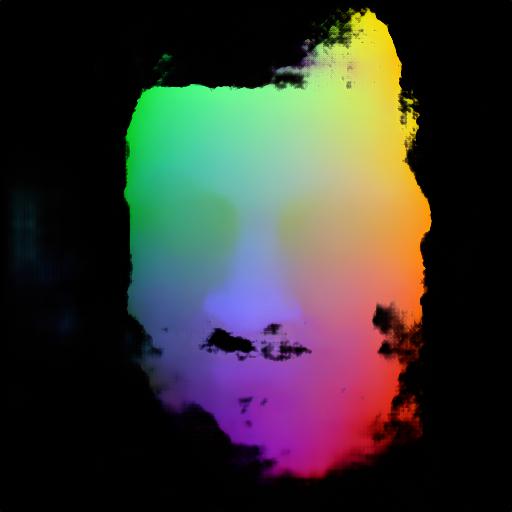} &
\includegraphics[width=0.092\textwidth]{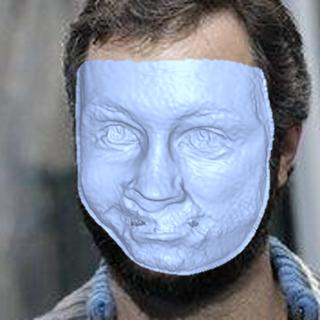} &

\includegraphics[width=0.092\textwidth]{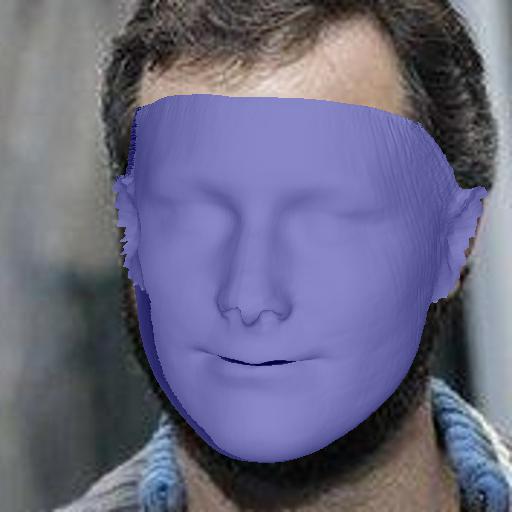} & 

\includegraphics[width=0.092\textwidth]{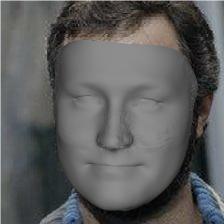} &
\includegraphics[width=0.092\textwidth]{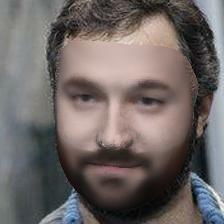} &

\includegraphics[width=0.092\textwidth]{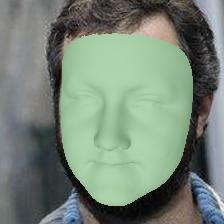} & 
\includegraphics[width=0.072\textwidth]{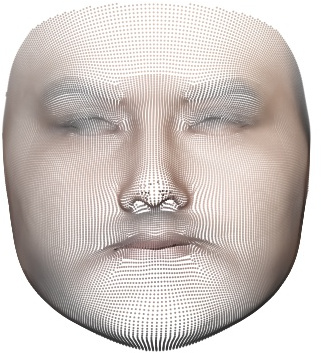} &
\includegraphics[width=0.092\textwidth]{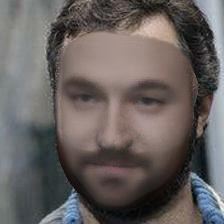} \\ 

\includegraphics[width=0.092\textwidth]{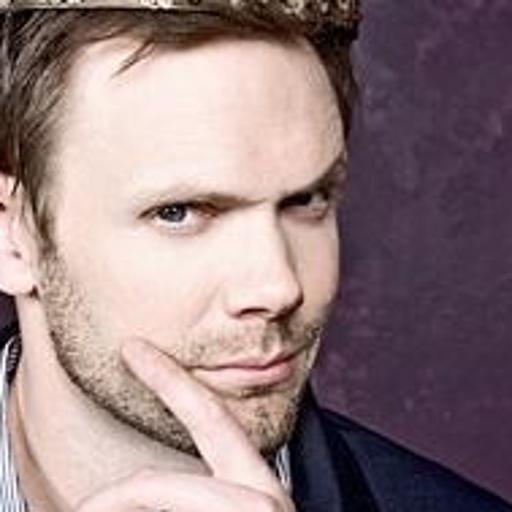} &

\includegraphics[width=0.092\textwidth]{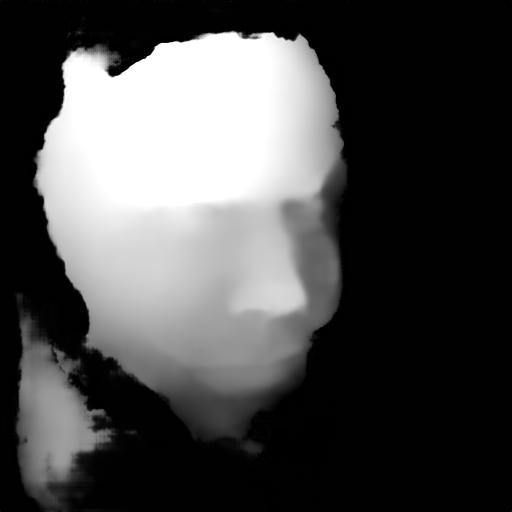} &
\includegraphics[width=0.092\textwidth]{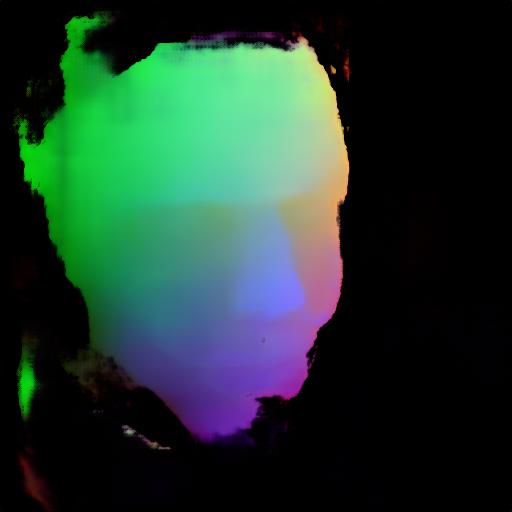} &
\includegraphics[width=0.092\textwidth]{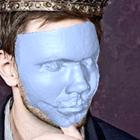} &

\includegraphics[width=0.092\textwidth]{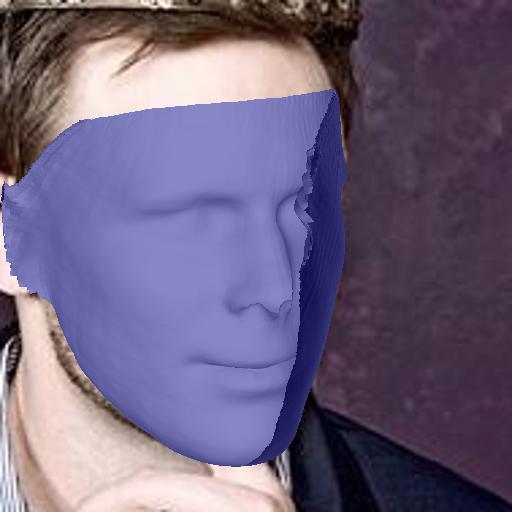} & 

\includegraphics[width=0.092\textwidth]{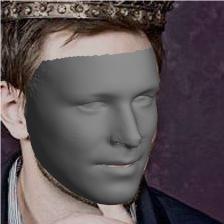} &
\includegraphics[width=0.092\textwidth]{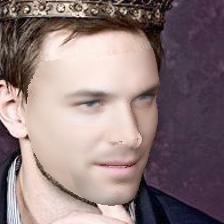} &

\includegraphics[width=0.092\textwidth]{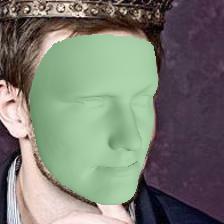} & 
\includegraphics[width=0.072\textwidth]{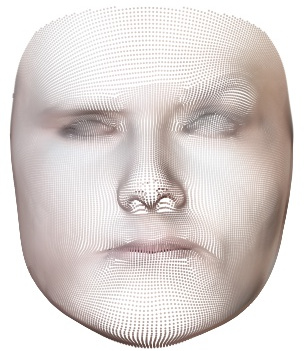} &
\includegraphics[width=0.092\textwidth]{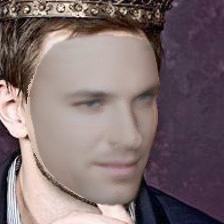}\\ 

\end{tabular}
\caption{3D reconstruction results compared to Sela \etal~\cite{sela2017unrestricted}. We show the estimated depth, correspondence map and shape for the method proposed by Sela \etal~\cite{sela2017unrestricted}, and we find occlusions can cause serious problems in their output maps.}
\label{fig:3drecon_sela}
\end{center}
\end{figure*}

\noindent\textbf{Comparison to Sela \etal \cite{sela2017unrestricted}.} 
Their elementary image-to-image network is trained on synthetic data generated by the linear model. Due to the domain gap between synthetic and real images, the network output tends to be unstable on some occluded regions for the in-the-wild testing (Fig.~\ref{fig:3drecon_sela}), which leads to failure in later steps. By contrast, our coloured mash decoder is trained on the real-world unconstrained dataset in an end-to-end self-supervised fashion, thus our model is robust in handling the in-the-wild variations. In addition, the method of Sela \etal~\cite{sela2017unrestricted} requires a slow off-line nonrigid registration step ($\sim180$s) to obtain a hole-free reconstruction from the predicted depth map. Nevertheless, the proposed coloured mesh decoder can run extremely fast. Furthermore, our method is complementary to Sela \etal~\cite{sela2017unrestricted}'s fine detail reconstruction module. Employing Shape from Shading (SFS) \cite{kemelmacher20113d} to refine our fitting results could lead to better results with details.

\begin{figure*}[t!]
\begin{center}
\small
\begin{tabular}{c@{\hskip 2mm}c@{\hskip 1mm}c@{\hskip 2mm}c@{\hskip 1mm}c@{\hskip 2mm}c@{\hskip 2mm}c@{\hskip 1mm}c@{\hskip 1mm}c@{}}

Input & \multicolumn{2}{c}{MoFA \cite{tewari2017mofa}} & PRNet \cite{feng2018joint} & \multicolumn{2}{c}{N-3DMM \cite{tran2018learning}} & \multicolumn{3}{c}{CMD} \\

\includegraphics[width=0.092\textwidth]{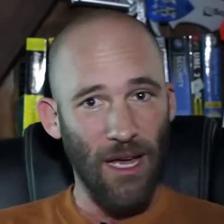} &

\includegraphics[width=0.092\textwidth]{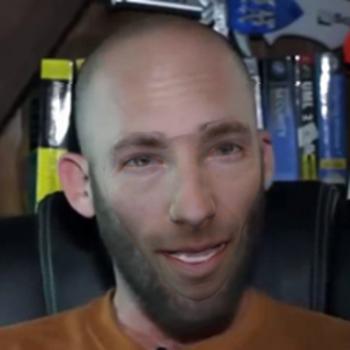} &
\includegraphics[width=0.092\textwidth]{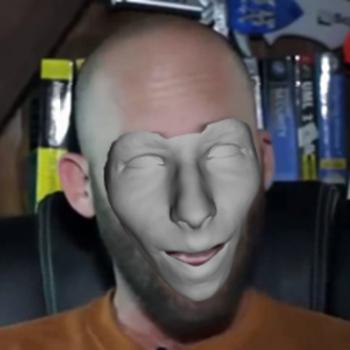}&

\includegraphics[width=0.092\textwidth]{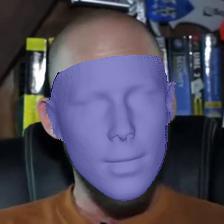} &

\includegraphics[width=0.092\textwidth]{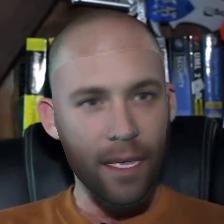} &
\includegraphics[width=0.092\textwidth]{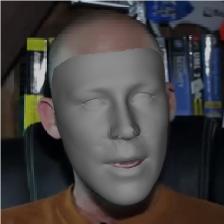} & 

\includegraphics[width=0.092\textwidth]{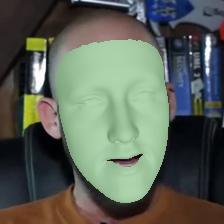} &  
\includegraphics[width=0.072\textwidth]{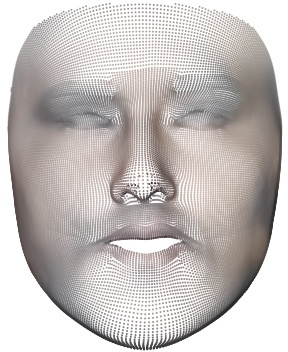} & 
\includegraphics[width=0.092\textwidth]{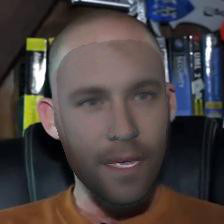} \\ 

\includegraphics[width=0.092\textwidth]{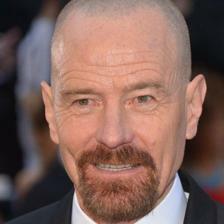} &

\includegraphics[width=0.092\textwidth]{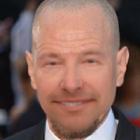} &
\includegraphics[width=0.092\textwidth]{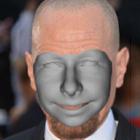} &

\includegraphics[width=0.092\textwidth]{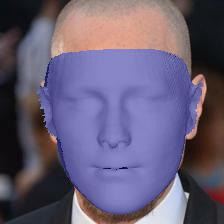} &

\includegraphics[width=0.092\textwidth]{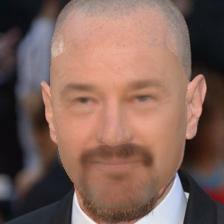} &
\includegraphics[width=0.092\textwidth]{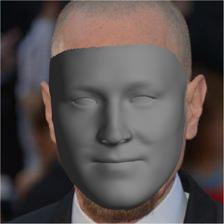} &

\includegraphics[width=0.092\textwidth]{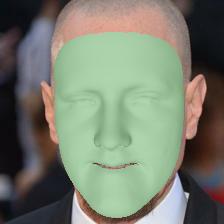} & 
\includegraphics[width=0.072\textwidth]{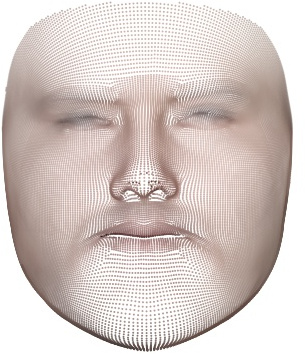} & 
\includegraphics[width=0.092\textwidth]{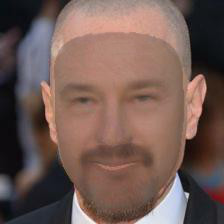} \\

\end{tabular}

\caption{3D face reconstruction results compared to MoFA~\cite{tewari2017mofa} on samples from the 300VW dataset~\cite{shen2015first} (first row) and the CelebA dataset~\cite{liu2015deep} (second row). The reconstructed shapes of MoFA suffer from unnatural surface deformations when dealing with challenging texture, \ie beard. By contrast, our non-linear coloured mesh decoder is more robust to these variations.}
\label{fig:3drecon_tewari}
\end{center}
\end{figure*}

\noindent\textbf{Comparison to MoFA~\cite{tewari2017mofa}.} 
The monocular 3D face reconstruction method, MoFA, proposed by Tewari \etal~\cite{tewari2017mofa}, employs an unsupervised fashion to learn 3DMM fitting in the wild. However, their reconstruction space is still limited to the linear bases.
Hence, their reconstructions suffer from unnatural surface deformations when dealing with very challenging texture,\ie beard, as shown in Fig.~\ref{fig:3drecon_tewari}. By contrast, our method employs a non-linear coloured mesh decoder to jointly reconstruct shape and texture. Therefore, our method can achieve high-quality reconstruction results even under hairy texture.

\begin{figure}[t!]
\begin{center}
\begin{tabular}{c@{\hskip 0.1mm}c@{\hskip 0.1mm}c@{\hskip 0.1mm}c@{\hskip 0.1mm}c}
Input &  VRN \cite{jackson2017large} & PRNet \cite{feng2018joint} & \small N-3DMM \cite{tran2018learning} & CMD \\

\includegraphics[width=0.081\textwidth]{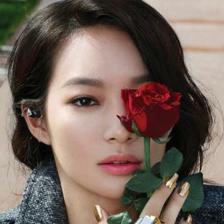} &
\includegraphics[width=0.081\textwidth]{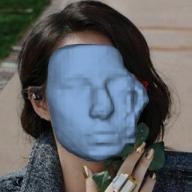} &
\includegraphics[width=0.081\textwidth]{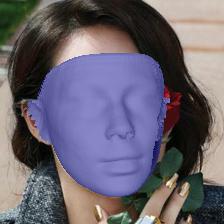} &
\includegraphics[width=0.081\textwidth]{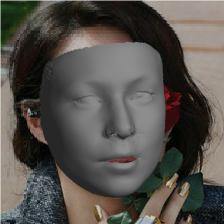} &
\includegraphics[width=0.081\textwidth]{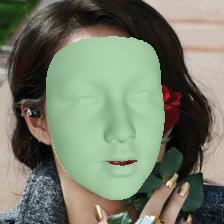} \\

\includegraphics[width=0.081\textwidth]{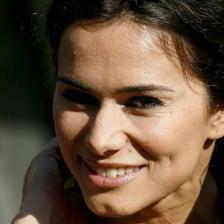} &
\includegraphics[width=0.081\textwidth]{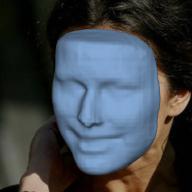} &
\includegraphics[width=0.081\textwidth]{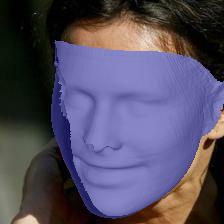} &
\includegraphics[width=0.081\textwidth]{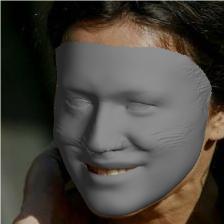} &
\includegraphics[width=0.081\textwidth]{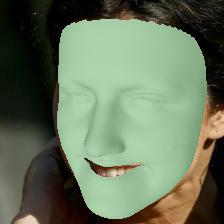} \\

\includegraphics[width=0.081\textwidth]{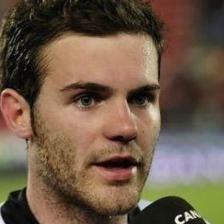} &
\includegraphics[width=0.081\textwidth]{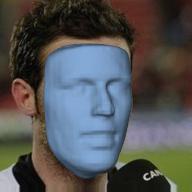} &
\includegraphics[width=0.081\textwidth]{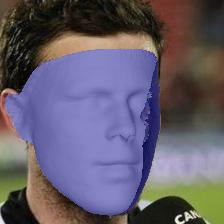} &
\includegraphics[width=0.081\textwidth]{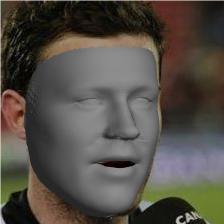} &
\includegraphics[width=0.081\textwidth]{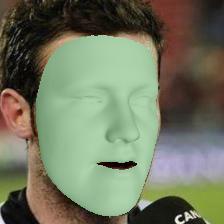} \\

\end{tabular}
\caption{3D reconstruction results compared to VRN \cite{jackson2017large} on the CelebA dataset~\cite{liu2015deep}. Volumetric shape representation results in non-smooth 3D shape and loses correspondence between reconstructed shapes. UV position map representation used in PRNet \cite{feng2018joint} and N-3DMM \cite{tran2018learning} has comparable performance with our method but the computation complexity is much higher and the model size is much larger.}
\label{fig:3drecon_jackson}
\end{center}
\end{figure}

\noindent\textbf{Comparison to VRN \cite{jackson2017large}.} We also compare our approach with a direct volumetric regression method proposed by Jackson \etal~\cite{jackson2017large}. VRN directly regresses a 3D shape volume via an encoder-decoder network with skip connection (\ie Hourglass structure) to avoid explicitly using a linear 3DMM prior. This strategy potentially helps the network to explore a larger solution space than the linear model. However, this method discards the correspondence between facial meshes and the regression target is very large in size. Fig.~\ref{fig:3drecon_jackson} shows a visual comparison of 3D face reconstructions between VRN and our method. In general, VRN can robustly handle in-the-wild texture variations. However, due to the volumetric shape representation, the surface is not smooth and does not preserve details. By contrast, our method directly models shape and texture of vertices, thus the model size is more compact and the output results are more smooth.

Besides qualitative comparisons with state-of-the-art 3D face reconstruction methods, we also conducted quantitative comparisons on the FaceWarehouse dataset~\cite{cao2014facewarehouse} and the Florence dataset~\cite{bagdanov2011florence} to show the superiority of the proposed coloured mesh decoder.

\begin{figure}
\begin{center}
\begin{tabular}{c@{\hskip 2mm}c}
\includegraphics[width=0.85\columnwidth]{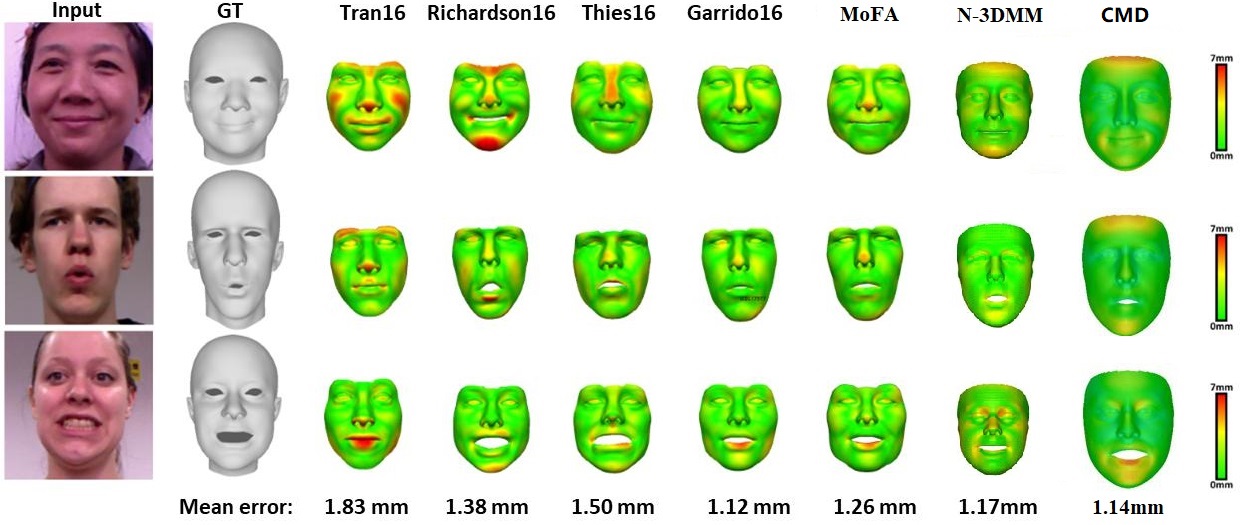}&
\includegraphics[width=0.09\columnwidth]{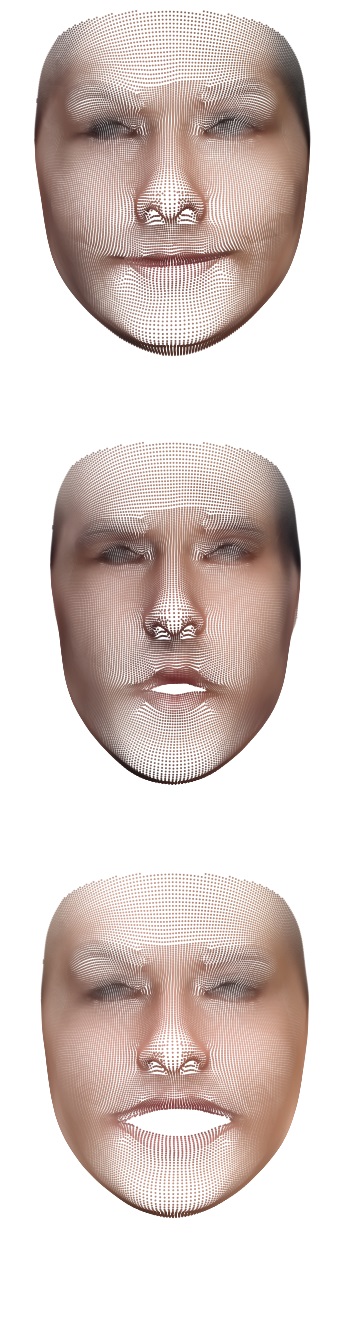}\\
\end{tabular}
\caption{Quantitative evaluation of 3D face reconstruction on the FaceWarehouse dataset~\cite{cao2014facewarehouse}. We achieved comparable performance compared to Garrido \etal~\cite{garrido2016reconstruction} and N-3DMM~\cite{tran2018learning}.}
\label{fig:3drecon_FaceWarehouse}
\end{center}
\end{figure}

\noindent\textbf{FaceWarehouse.} Following the same setting in~\cite{tewari2017mofa,tran2018learning}, we also quantitatively compared our method with prior works on 9 subjects from the FaceWarehouse dataset~\cite{cao2014facewarehouse}. Visual and quantitative comparisons are illustrated in Fig.~\ref{fig:3drecon_FaceWarehouse}. We achieved comparable results with Garrido \etal~\cite{garrido2016reconstruction} and N-3DMM \cite{tran2018learning}, while surpassing all other regression methods \cite{tran2017regressing,richardson2017learning,tewari2017mofa}. As shown on the right side of Fig.~\ref{fig:3drecon_FaceWarehouse}, we can easily infer the expression of these three samples from their coloured vertices.

\begin{figure}[h!]
\centering
\subfigure[CED Curves]{
\label{fig:florenceced}
\includegraphics[width=0.22\textwidth]{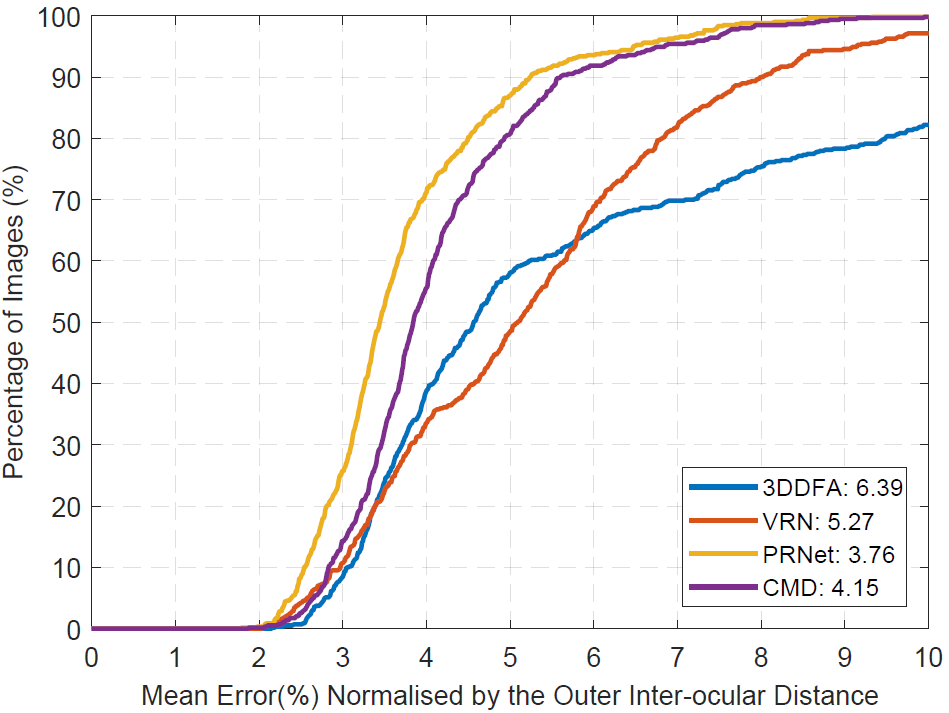}}
\subfigure[Pose-specific NME]{
\label{fig:florencenme}
\includegraphics[width=0.22\textwidth]{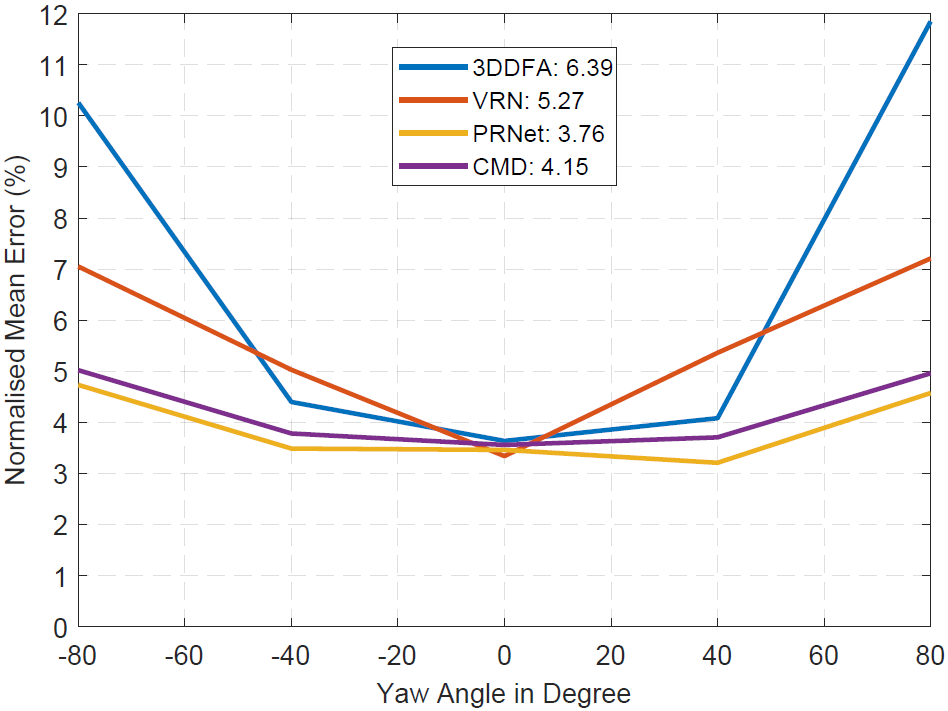}}
\caption{3D face reconstruction results on the Florence dataset~\cite{bagdanov2011florence}. The Normalized Mean Error of each method is showed in the legend.}
\label{fig:florence}
\end{figure}

\noindent\textbf{Florence.} Following the same setting in~\cite{jackson2017large,feng2018joint}, we also quantitatively compared our approach with state-of-the-art methods (\eg VRN \cite{jackson2017large} and PRNet \cite{feng2018joint}) on the Florence dataset~\cite{bagdanov2011florence}. The face bounding boxes were calculated from the ground truth point cloud and the face images were cropped and used as the network input. Each subject was rendered with different poses as in \cite{jackson2017large,feng2018joint}: pitch rotations of $-15^{\circ}$, $20^{\circ}$ and $25^{\circ}$ and raw rotations between $-80^{\circ}$ and $80^{\circ}$. We only chose the common face region to compare the performance. For evaluation, we first used the Iterative Closest Points (ICP) algorithm to find the corresponding nearest points between our model output and ground truth point cloud and then calculated Mean Squared Error (MSE) normalized by the inter-ocular distance of 3D coordinates.

Fig.~\ref{fig:florenceced} shows that our method obtained comparable results with PRNet. To better evaluate the reconstruction performance of our method across different poses, we calculated the NME under different yaw angles. As shown in Fig.~\ref{fig:florencenme}, all the methods obtain good performance under the near frontal view. However, 3DDFA and VRN fail to keep low error as the yaw angle increases. The performance of our method is relatively stable under pose variations and comparable with the performance of PRNet under profile views.

\subsection{Running Time and Model Size Comparisons}

\begin{table}[t!]
\begin{center}
\begin{tabular}{c|c c||c c}
\hline
 & \multicolumn{2}{c||}{Time} & \multicolumn{2}{c}{Size} \\
\hline
Method & E & D & E & D \\ 
\hline
Sela \etal~\cite{sela2017unrestricted} & \multicolumn{2}{c||}{$10$ ms } & \multicolumn{2}{c}{$ 1.2$G } \\
VRN \cite{jackson2017large}     & \multicolumn{2}{c||}{$10$ ms } & \multicolumn{2}{c}{$ 1.5$G } \\
PRNet \cite{feng2018joint}      & \multicolumn{2}{c||}{$10$ ms } & \multicolumn{2}{c}{$ 153$M } \\
MoFA \cite{tewari2017mofa}      & $ 4$ms & $1.5$ms & $100$M & $120$M \\  
N-3DMM \cite{tran2018learning}  & $ 2.7$ms  & $ 5.5$ ms & $ 76$M & $ 76$M \\ 
\hline
PCA Shape                 & $ 1.5$ms & $ 1.5$ms & \multicolumn{2}{c}{$ 129$M } \\
PCA Texture            & $ 1.7$ms & $ 1.7$ms & \multicolumn{2}{c}{$ 148$M } \\
\hline
CMD ($f_{SA}$=256)       & $ 2.7$ms  & ${\bf 0.367}$ms    & $ 76$M  & ${\bf 17}$M \\
\hline
\end{tabular}
\end{center}
\caption{Running time and model size comparisons of various 3D face reconstruction methods. Our coloured mesh decoder can run at $0.367$ms on CPU with a compact model size of $17$MB.} 
\label{tab:abl_runtime}
\end{table}

In Tab.~\ref{tab:abl_runtime}, we compare the running time and the model size for multiple 3D reconstruction approaches. Since some methods were not publicly available~\cite{sela2017unrestricted,tewari2017mofa,tran2018learning}, we only provide an approximate estimation for them. Sela \etal~\cite{sela2017unrestricted}, VRN~\cite{jackson2017large} and PRNet~\cite{feng2018joint} all use an encoder-decoder network with similar running time. However, Sela \etal~\cite{sela2017unrestricted} requires an expensive nonrigid registration step as well as a refinement module. 

Our method gets a comparable encoder running time with N-3DMM~\cite{tran2018learning} and MoFA~\cite{tewari2017mofa}. However, N-3DMM~\cite{tran2018learning} requires decoding features via two CNNs for shape and texture, respectively. MoFA~\cite{tewari2017mofa} directly uses liner bases, and the decoding step is a single multiplication around $1.5$ms for 28K points. By contrast, the proposed coloured mesh decoder only needs one efficient mesh convolution network. On CPU (Intel i9-7900X@3.30GHz), our method can complete coloured mesh decoding within 0.367 ms (2500FPS), which is even faster than using linear shape bases. The model size of our non-linear coloured mesh decoder ($17$M) is almost one-seventh of the liner shape bases ($120$MB) employed in MoFA. Most importantly, the capacity of our non-linear mesh decoder is much higher than that of the linear bases as proved in the above experiments.

\section{Conclusions}

In this paper, we presented a novel non-linear 3DMM method using mesh convolutions. Our method decodes both shape and texture directly on the mesh domain with compact model size ($17$MB) and very low computational complexity (over 2500 FPS on CPU). Based on the mesh decoder, we propose an image encoder plus a coloured mesh decoder structure that reconstruct the texture and shape directly from an in-the-wild 2D facial image. Extensive qualitative visualization and quantitative reconstruction results confirm the effectiveness of the proposed method.
 
\section{Acknowledgements}

Stefanos Zafeiriou acknowledges support from EPSRC Fellowship DEFORM (EP/S010203/1) and a Google Faculty Fellowship. Jiankang Deng acknowledges insightful advice from friends (\eg Sarah Parisot, Yao Feng, Luan Tran and Grigorios Chrysos), financial support from the Imperial President's PhD Scholarship, and GPU donations from NVIDIA.

{\small
\bibliographystyle{ieee}
\bibliography{egbib}
}

\end{document}